%% file: spectra_vla_tpami_manuscript.tex
\documentclass[10pt,journal,compsoc]{IEEEtran}
\usepackage{amsmath,amssymb}
\usepackage{booktabs}
\usepackage{graphicx}
\usepackage{multirow}
\usepackage{url}
\usepackage[table]{xcolor}
\ifCLASSOPTIONcompsoc
  \usepackage[nocompress]{cite}
\else
  \usepackage{cite}
\fi
\usepackage[hidelinks]{hyperref}
\graphicspath{{figures/}}

\definecolor{proposedgray}{gray}{0.94}
\title{SPECTRA-VLA: Base-Fine Hyperspectral Token Transfer at Frozen Vision-Language-Action Interfaces}
\author{Chih-Chung Hsu\thanks{Chih-Chung Hsu is with National Yang Ming Chiao Tung University, Taiwan (e-mail: chihchung@nycu.edu.tw).}}
\begin{document}
\IEEEtitleabstractindextext{%
\begin{abstract}
Hyperspectral imaging distinguishes RGB-similar objects, but a pretrained vision-language-action model needs a state its frozen decoder can use. SPECTRA-VLA provides a native-width spectral upgrade without retraining the visual encoder or action decoder. A learned continuous residual enriches each target token through its own normalized false-color reference; projection, norm control and validation-calibrated injection regulate the change. Across 86 held-out acquisitions, category balanced accuracy rises from 0.812 to 0.883. With identical residuals and scale, own-token addition improves reuse of an external category-derived probe on frozen decoder logits: 0.701 versus 0.590 with pooled replacement, at similar image-disjoint recognition. Five grouped tests covering the same 86 acquisitions recover within-category spectral shape with $R^2=0.802/0.337$ from continuous and posterior-code states. Category-matched spectral swaps further show that own-token composition approximately doubles the target-specific recovery increment after frozen decoding. Spectral enrichment and receiving-token preservation thus work together at the pretrained interface.
\end{abstract}
}
\maketitle
\IEEEdisplaynontitleabstractindextext

\IEEEraisesectionheading{\section{Introduction}}
\IEEEPARstart{V}{ision-language-action} (VLA) models connect pretrained visual-language representations to robot action tokens \cite{rt1,rt2,openvla,spatialvla}. Their RGB-optimized interfaces can miss distinctions needed for waste-object recognition: cross-category SpectralWaste objects have similar paired true-RGB VLA tokens but different spectra. Hyperspectral imaging (HSI) supplies additional measurements, but replacing the visual representation can discard the token structure that the pretrained decoder expects. We address this upgrade at the existing encoder--decoder connection, retaining both pretrained components.

The central question is how to add spectral discrimination while preserving token-specific content of the normalized reference grid. SPECTRA-VLA couples continuous spectral enrichment with receiving-token preservation: a learned residual enters through each target token's own reference, at the native token width and a validation-calibrated scale. Category probabilities compress within-category spectral variation; pooled replacement omits each receiving token's own reference from the inserted branch. Matched content, composition and movement comparisons establish the distinct benefits of retaining both the spectrum and the receiving grid.

Our contributions are:
\begin{itemize}
\item \textbf{Receiving-token-preserving spectral transfer.} A continuous spectral residual enters through each token's own reference, retaining the pretrained encoder, action decoder and native token width. Across 86 held-out acquisitions, category bAcc improves from $0.812$ to $0.883$, with paired session gain $0.071\ [0.050,0.107]$. A second backbone reproduces fixed-split enrichment ($0.812$ to $0.914$).
\item \textbf{Continuous enrichment beyond category compression.} Across five session-grouped tests, continuous and posterior-code states recover within-category spectral shape with $R^2=0.802/0.337$ under the same composition and allowance. Learning the residual also improves category bAcc over a label-free map ($0.865/0.773$) while reducing mean decoder KL ($0.048/0.062$) in the matched 16-session comparison.
\item \textbf{Spectral-content--composition interaction.} Across 86 acquisitions, own-token addition approximately doubles target-specific recovery increments after frozen decoding (interaction $0.0171\ [0.0121,0.0228]$; Table~\ref{tab:same-residual-decoder}d). Complementary image-disjoint tests show $0.111$ better reference-fixed reuse than pooled replacement at similar recognition. At a common allowance, base-fine matches the unprojected ablation's rounded recognition with smaller disturbance tails.
\end{itemize}

\begin{figure*}[t]
\centering
\includegraphics[width=\textwidth]{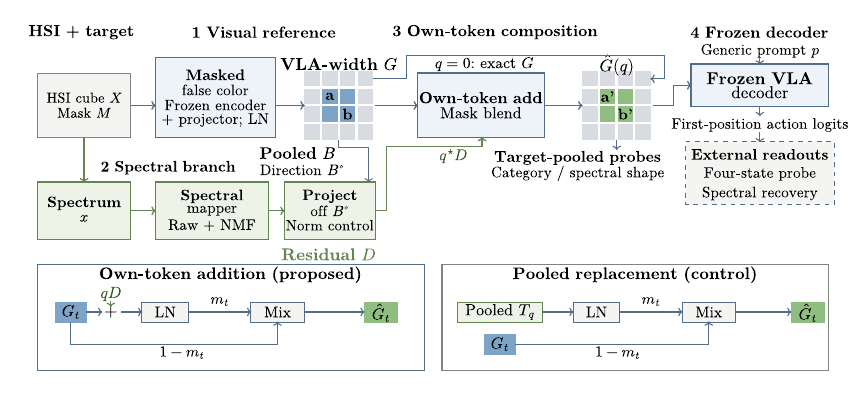}
\caption{SPECTRA-VLA adds spectral residuals after the frozen visual backbone and pretrained projector. LN defines native-width reference $G$. Lower panels share $D,q,m_t$ and skip $(1-m_t)G_t$. For $q>0$, pooled $T_q=\mathrm{LN}_{\epsilon}(B^{\circ}+qD)$; only addition retains $G_t$ inside the normalized branch. Pooled-token probes measure category and spectral shape. The frozen decoder consumes the full grid and generic prompt $p$; external first-position readouts measure four-state information and spectral recovery. Category labels train the mapper; validation selects $q$. Projection and norm control constrain the residual. Non-target tokens are unchanged; $q=0$ exactly returns $G=\mathrm{LN}(Z_{\mathrm{base}})$. Blue/green identifies reference/spectral paths.}
\label{fig:architecture}
\end{figure*}

Category supervision learns discriminative continuous content \cite{spectralwaste}; own-token composition preserves the receiving grid; projection defines a base-complement direction; norm and validation calibration control the executed disturbance (Fig.~\ref{fig:architecture}). Category recognition and spectral recovery measure enrichment. External four-state decoder-logit probes distinguish reuse of a reference-trained decision function from accessibility after refitting.

\section{Related Work}
\subsection{VLA Models and Beyond-RGB Sensing}
OpenVLA is an open 7B-parameter action-token model trained on 970k Open X-Embodiment episodes; SpatialVLA adds spatial representations learned from 1.1 million episodes \cite{openvla,spatialvla}. PointVLA injects point clouds into action-expert blocks, and PointACT trains an action expert over hierarchical point-cloud features \cite{pointvla,pointact}. HARP-VLA preserves discrimination while aligning human and robot videos \cite{harpvla}. These methods use video or robot-action supervision.

OmniVLA and MuseVLA provide the closest beyond-RGB action comparison \cite{omnivla,musevla}. Both convert thermal, acoustic, or radar measurements into spatially grounded sensor images and train downstream VLA components with demonstrations. OmniVLA freezes its vision encoder; MuseVLA additionally selects sensors as task-conditioned tools. Tactile VLAs condition visual features, modulate tactile tokens, distill tactile-aware behavior, or overlay tactile fields for action generation \cite{tacfilm,tacvla,hapticvla,tapvla}. SPECTRA-VLA holds both encoder and action decoder fixed. This unchanged downstream consumer motivates our separate measurements of recognition, decoder disturbance, and retention of an established readout.

\subsection{Spectral Representations and RGB Transfer}
SpectralGPT, HyperSIGMA, DOFA, HyperSL, MaskedSST, and HyperFree learn spectral-image representations across remote-sensing domains \cite{spectralgpt,hypersigma,dofa,hypersl,maskedsst,hyperfree}; HyperVision covers heterogeneous ground-based sensors \cite{hypervision}. HSI-Adapter mixes original and injected ViT tokens through spectral--spatial interaction and trains a segmentation decoder \cite{hsiadapter}. HyA-T freezes SeqTrack's original encoder and autoregressive decoder, learning internal attention and feedforward adapters with tracking supervision \cite{hyat}. SPECTRA-VLA inserts at the existing visual-token connection: category supervision learns continuous spectral content, while composition preserves each receiving token's reference. UniDiff anchors diffusion-based HSI/SAR classification with pseudo-RGB \cite{unidiff}.

Low-rank updates, bottleneck adapters, and visual prompts provide compact pretrained-model adaptation \cite{lora,adaptformer,vpt}. BLIP-2 places a learned querying transformer between frozen image and language models \cite{blip2}; cross-modal backward-compatible training aligns new and old vision-language embeddings for retrieval \cite{xbt}. Orthogonal Residual Update projects updates off an internal residual stream during training \cite{oru}. ORCA uses a pretrained fundus-image anchor, orthogonal cross-modal residuals and norm control for choroidal-thickness regression \cite{orca}. Feature-wise concatenation expands feature width. TokenFusion replaces low-importance tokens while retaining position through residual positional alignment \cite{tokenfusion}; our branch retains token content $G_t$, not only position. BCAF learns an RGB--HSI segmentation endpoint on SpectralWaste \cite{bcaf}.

SPECTRA-VLA learns a category-supervised continuous spectral residual at the native-width post-projector interface. Equation~\eqref{eq:composition} combines it with each receiving token's own content before normalization; the unchanged decoder consumes the full grid. Category-code and residual-placement controls test content and composition (Tables~\ref{tab:spectral-shape}, \ref{tab:same-residual-decoder}); category-matched swaps test their interaction, and common-constraint calibration measures recognition and disturbance (Table~\ref{tab:common-budget-main}).

\section{Problem Setting and Evidence Scope}
An observation consists of HSI cube $X_i$, supplied target mask $M_i$, and a generic prompt without a category name. The target spectrum is $x_i=\operatorname{pool}(X_i,M_i)$. A frozen VLA exposes visual states $E(I_i)$ and a decoder $A(Z_i,p_i)$. Our bridge composes an HSI-conditioned state at the original visual-token width and feeds it to $A$ without changing VLA parameters.

The reference is a fixed HSI-derived SWIR false-color view, not paired true RGB. Paired RGB is used separately to identify a stress set whose cross-category objects have similar RGB-VLA tokens. SpectralWaste's six labels (film, basket, cardboard, video tape, filament, and bag) are operational object categories rather than verified physical-composition identities \cite{spectralwaste}. A four-state target is fixed by mapping cardboard to 0; film, filament, and bag to 1; basket to 2; and video tape to 3. This \emph{category-derived proxy state} is a controlled coarsening of category information, not an independently labeled robot action.

\emph{Pooled-token category} and \emph{pooled-token proxy} readouts classify pooled states. The \emph{decoder-logit probe} (``decoder probe'' in tables) is an external four-state ridge classifier on first-position (FP) frozen action-symbol logits. A \emph{refitted decoder-logit probe} learns separately for each representation, measuring accessible information; a \emph{reference-fixed decoder-logit probe} retains the reference-trained standardizer and classifier, measuring decision-function reuse. Four-state content interventions instead retain the \emph{correct-spectrum-trained probe}. In decoder tables, ``refit'' means refitted logit bAcc; ``base-fixed FP,'' ``Fixed bAcc'' and ``Ref.-fixed'' mean reference-fixed logit bAcc. ``C-proxy'' identifies the category-derived target, with the table heading specifying pooled tokens or decoder logits. Native width, non-target preservation and zero-injection identity follow by construction; recognition, spectral recovery and decoder behavior are measured.

\section{Base-Fine Hyperspectral Transfer}
\label{sec:base-fine}
SPECTRA-VLA trains a spectral branch and auxiliary heads while both VLA components remain frozen.\footnote{Implementation assistance from OpenAI Codex and ChatGPT~\cite{openai_codex,openai_chatgpt} is disclosed in the Acknowledgment.} The pooled reference controls residual direction and norm; the mask locates insertion through each receiving token's own content. Validation selects injection scale, and the training heads are discarded.

\subsection{False-Color Reference and Target Support}
The sensor spans 900--1700~nm. Fixed bands $[119,60,0]$ with masked exterior pixels pass through the frozen visual backbone and its pretrained visual-to-language projector $E$, yielding post-projector embeddings $Z_i^{\mathrm{base}}\in\mathbb{R}^{256\times4096}$. The bridge acts here, before the frozen decoder. Area-pooling $M_i$ onto the $16\times16$ grid gives occupancy $\widetilde m_{it}$. We use
\begin{equation}
m_{it}=\frac{\widetilde m_{it}}{\max_t\widetilde m_{it}+\delta},\qquad\delta=10^{-6},
\label{eq:mask-weight}
\end{equation}
which preserves relative occupancy and gives nonempty small objects approximately unit peak weight. Non-affine LayerNorm defines the reference grid and pooled state:
\begin{align}
G_{it}&=\mathrm{LN}_{\epsilon}(Z_{it}^{\mathrm{base}}),\qquad\epsilon=10^{-5},\\
B_i&=\mathrm{LN}_{\epsilon}\!\left(\frac{\sum_t m_{it}G_{it}}{\sum_t m_{it}+10^{-6}}\right).
\label{eq:base}
\end{align}
The operational reference $G$ provides a common token scale for base-relative geometry and executed-movement calibration. Normalization deliberately changes the interface while model parameters remain frozen: the cached-runtime raw-to-$G$ diagnostic gives mean action KL $2.610$ and top-1 agreement $0.163$ (Supplement S7.2). Preservation comparisons begin at $G$, which zero injection exactly restores. Projection and scaling use $B_i^{\circ}=\mathrm{LN}_{\epsilon}(B_i)$.

\subsection{Spectral Fine Residual}
\label{sec:spectral-fine}
The fine branch pairs the target's 120 raw bands with 16 NMF coordinates. The mapper learns category-discriminative information; projection and rescaling define its base-relative insertion. Standardization, percentile scaling and NMF are fitted only on the relevant training partition:
\begin{equation}
s_i=[\operatorname{Std}(x_i);\operatorname{Std}(a_i)].
\end{equation}
A bottleneck with hidden width $r$ produces
\begin{equation}
u_i=W_{\mathrm{up}}\operatorname{GELU}\!\left(W_{\mathrm{down}}\mathrm{LN}_{\mathrm{aff}}(s_i)\right).
\label{eq:bottleneck}
\end{equation}
Its output becomes base-dependent through
\begin{align}
\widetilde D_i&=u_i-\frac{\langle u_i,B_i^{\circ}\rangle}{\|B_i^{\circ}\|_2^2+\delta}B_i^{\circ},\label{eq:projection}\\
D_i&=\rho\|B_i^{\circ}\|_2\frac{\widetilde D_i}{\|\widetilde D_i\|_2+\delta},\qquad\rho=0.1.
\label{eq:residual}
\end{align}
Projection defines an approximately base-orthogonal spectral direction for the geometric analysis and matched interventions. Equation~\eqref{eq:residual} rescales nondegenerate residuals toward $\rho\|B_i^{\circ}\|_2$. For finite positive $\delta$, $\|qD_i\|_2\leq q\rho\|B_i^{\circ}\|_2$ in real arithmetic. At the largest searched scale $q=4$, the pre-composition residual budget is $0.4$ of the base norm. Recognition and decoder divergence are measured separately from this geometric constraint.

Training operates on the pooled-vector state $T_q=\mathrm{LN}_{\epsilon}(B_i^{\circ}+qD_i)$ for $q>0$, with $T_0=B_i^{\circ}$. Each item draws $q_i\sim\mathcal U(0,4)$. Here $T_q$ is the pooled training surrogate; evaluation instead uses the composed grid and its pool $C_q$ defined below. Category supervision minimizes
\[
\begin{aligned}
\mathcal L={}&\operatorname{CE}_w(h_c(T_{q_i}),y_i)
+\lambda_f\operatorname{CE}_w(h_f(D_i),y_i)\\
&+0.1\operatorname{CE}_w(h_c(T_0),y_i)
+10^{-4}\operatorname{mean}(D_i^2).
\end{aligned}
\]
The fine-loss weight $\lambda_f$ is 1 in fixed-split mechanism studies and inner-selected in principal session holdout.
The shared head $h_c$ learns to classify both the reference and states at sampled injection scales; the separate head $h_f$ supervises the residual itself. The reference term supplies no adapter gradient because $T_0$ bypasses the residual. For training class counts $n_c>0$ over $C$ classes, $w_c=C/(n_c\sum_{k=1}^{C}n_k^{-1})$ and $\mathrm{CE}_w=-w_y\log p_y$. Each classification term averages over minibatch $\mathcal B$ as $\sum_{i\in\mathcal B}\omega_i\mathrm{CE}_{w,i}/\sum_{i\in\mathcal B}\omega_i$, with $\omega_i=4$ on training stress-pair endpoints and $1$ otherwise. Matched controls share this reduction except the no-pair-weighting ablation. The denominator excludes class weights; the residual penalty is an unweighted mean. Equation~\eqref{eq:residual}, rather than the nearly constant penalty on rescaled nondegenerate residuals, controls the norm. Training heads are discarded; only category labels supervise the adapter.

\subsection{Composition, Endpoint, and Geometry}
The controlled residual is broadcast into target tokens:
\begin{equation}
\widehat G_{it}(q)=\begin{cases}
G_{it},&q=0,\\
(1-m_{it})G_{it}+m_{it}\mathrm{LN}_{\epsilon}(G_{it}+qD_i),&q>0.
\end{cases}
\label{eq:composition}
\end{equation}
Zero-weight tokens remain unchanged; the zero-scale branch returns $G$ without residual computation or normalization. Evaluation pools the composed grid into $C_q=\mathrm{LN}_{\epsilon}(\operatorname{pool}_m\widehat G_i(q))$; the decoder receives the complete grid. Per-token normalization and fractional-mask mixing distinguish $C_q$ from training state $T_q$.

\textbf{Executed insertion.} For mean-centered $G_t$ and $D_c=D-\bar D\mathbf1$ in real arithmetic, Eq.~\eqref{eq:composition} gives $\widehat G_t=a_tG_t+b_tD_c$ for $q>0$, with $a_t=1-m_t+m_tc_t$, $b_t=m_tc_tq$, and $c_t=(d^{-1}\|G_t+qD_c\|_2^2+\epsilon)^{-1/2}$. Occupancy and token-dependent normalization therefore determine how the shared payload enters each position. Addition uses the token's own reference in both the injected and skip branches; pooled replacement uses it only in the fractional skip. Table~\ref{tab:same-residual-decoder} isolates this choice with identical residuals; Supplement S7.1 gives its angular analysis.

For the idealized pre-normalization state $Y_q=B^{\circ}+qD$, setting $\delta=0$ and assuming a nondegenerate residual gives
\begin{align}
\cos(B^{\circ},Y_q)&=(1+\rho^2q^2)^{-1/2},\label{eq:preln-cosine}\\
w^\top Y_q&=w^\top B^{\circ}+q\,w^\top D.\label{eq:preln-geometry}
\end{align}
These identities describe pre-normalization angular control and directional margin changes; stabilizers, token normalization and fractional masks determine executed movement. Equal-mean calibration sets each base-fine seed to a preset validation mean-displacement target; final cosine and decoder KL are measured. At epoch 60, fixed-test, three-seed $T_q/C_q$ mean cosine is $0.9977$, relative distance $0.0569$, and residual coordinate-mean energy $0.0098\%$. Validation-selected checkpoints give $2.82\%$ disagreement between auxiliary-head $T_q/C_q$ predictions. The supplement expands occupancy-stratified margins and composition interventions.

\textbf{Selecting the executed state.} Define displacement $d_i(q)=\|C_i(q)-B_i\|_2/(\|B_i\|_2+10^{-12})$ and its validation mean $\bar d_{\mathrm{val}}(q)$. At a fixed trained checkpoint, a common allowance $b$ selects
\[
q_b^\star\in\arg\max_{q\in\mathcal Q}U_{\mathrm{val}}(q)
\quad\text{subject to}\quad\bar d_{\mathrm{val}}(q)\leq b,
\]
where $U=\mathrm{bAcc}+0.25\,\mathrm{Pair}$. Unrestricted selection omits the constraint. Equal-mean calibration instead calibrates each base-fine seed to a preset validation displacement; information matching targets a preset validation spectral $R^2$. These criteria select executed states, separately from training loss and the pre-normalization budget $q\rho$. Section~5.2 and Supplement S4 give each study's grids and ties; the principal 86-session selection remains distinct.

\section{Experimental Protocol}
\noindent\textit{AI assistance.} OpenAI Codex and ChatGPT~\cite{openai_codex,openai_chatgpt} assisted with experimental implementation and analysis, as disclosed in the Acknowledgment.
\subsection{Data and Split Boundaries}
\label{sec:data-preprocessing}
The released SpectralWaste corpus has 514/167/171 training/validation/test source images \cite{spectralwaste}. Four-connected target components of at least 100 resized pixels yield 2,295 objects, split 1,392/442/461. Cubes are linearly resampled from 224 to 120 bands and resized to $224^2$; raw object-mean spectra are retained without per-image radiometric normalization or object-wise unit scaling. The fixed sensor rendering uses $[119,60,0]$ with the released 16-bit scale, and image pixels outside the target are zeroed.

The fixed split is image-disjoint but not session-disjoint. The principal evaluation holds out each of 86 acquisition sessions containing 2,291 objects; four objects from two validation-only sessions are excluded. Every fitted spectral transform, NMF, candidate setting, pair threshold and readout is estimated inside outer training. Deterministic frozen visual tokens are precomputed without fitted spectral statistics. Selection uses 12 inner-training epochs and refitting uses 60 epochs. Separate fixed-coordinate diagnostics use a shorter 5/15 schedule and are reported as distinct estimands.

\subsection{Comparison Controls and Readouts}
\label{sec:matched-representations}
Controls distinguish spectral input, replacement, expanded width and residual constraints. The spectrum-only retargeter uses the bottleneck in Eq.~\eqref{eq:bottleneck}, category loss and coordinate-mean alignment to $B^{\circ}$. The high-anchor direct control additionally receives the base and retains an identity skip; validation requires cosine at least $0.94$ before utility selection. Supplement S1 and S3 give both objectives, capacities and grids. The fixed-map replacement maps 120 bands and 16 NMF coordinates to 4096 dimensions by a label-free isometry followed by row LayerNorm (Supplement S3.2). A replacement vector $v_i$ enters the full grid through
\begin{equation}
R_{it}(v_i)=(1-m_{it})G_{it}+m_{it}\mathrm{LN}_{\epsilon}(v_i),
\label{eq:replacement-grid}
\end{equation}
whereas additive controls use Eq.~\eqref{eq:composition} with their residual in place of $D_i$. The pooled-replacement control in Table~\ref{tab:same-residual-decoder} inserts the pooled base-plus-residual state $v_i=T_q$, using the same spectral residual and injection scale as own-token addition; its no-fine control inserts $v_i=B_i^{\circ}$. All preserve non-target tokens. Replacement removes the own-token reference from its injected branch but retains the fractional skip; addition uses it in both branches. For equal weights $m$ and a shared replacement vector, $R_t(v)-R_u(v)=(1-m)(G_t-G_u)$: differences shrink, collapsing only at $m=1$. Fine-only normalizes the proposed residual for replacement. Supplement S1 and S3 define the complete controls.

In the fixed-split comparison, base-fine and the base-conditioned direct mapper use hidden widths 64/32 and 271,184/275,024 mapper parameters; auxiliary heads explain their total-parameter difference. Fixed-split/mechanism controls share 60 epochs, AdamW, three seeds, preprocessing and category supervision; principal session holdout uses fold-specific seeds. Category and pooled-token proxy readouts use 256-D signed pooling, train-only standardization and class-balanced ridge. Validation category bAcc selects regularization; pair resolution requires both endpoints correct. Representation selection uses $U=\mathrm{bAcc}+0.25\,\mathrm{Pair}$ before fitting proxy readouts. Supplement S1 specifies pooling, grids and deterministic ties.

Mechanism studies train with independent $q_i\sim\mathcal U(0,4)$; minimum class-weighted validation CE of $h_c(T_1)$ selects a checkpoint within 60 epochs, excluding auxiliary and stress-pair terms. Unrestricted and allowance-constrained scales maximize validation utility; matched-movement and recovery studies target validation displacement or spectral recovery (Section~4.3). Refitted decoder-logit probes learn from each representation's logits; reference-fixed probes retain the reference-trained standardizer and classifier. Decoding receives the full grid; token readouts pool $C_q$. Fixed-epoch sensitivity uses epoch 60, rank 64 and $q=(2,4,4)$ for seeds $(0,1,2)$. Supplement S1/S4 gives regularization and fold choices. Test P95 values average per-seed computations.

\subsection{Frozen Decoder and Uncertainty}
OpenVLA receives the full 256-token grid and a generic sorting prompt \cite{openvla}. Its 32,064-coordinate runtime output supplies 256 first-position action symbols (IDs 31,744--31,999) to the four-state decoder-logit probes defined in Section~3. KL renormalizes this support, directed from reference to altered state.

Matched decoder comparisons jointly recompute reference and altered states in one inference configuration: bfloat16, batch four, no key--value cache, PyTorch 2.6.0, CUDA 12.4, Transformers 4.40.1 and scaled dot-product attention. Table~\ref{tab:common-budget-main} also uses the common category ridge grid. Earlier-runtime results form a separate sensitivity study (Supplement S1.6); captions identify the configuration and reference score of each comparison.

Raw-token reinsertion matches the official forward exactly on 171 RGB observations (Supplement S7.3). Section~4.1 reports the cached-runtime normalization diagnostic. SpatialVLA uses all 8,194 action coordinates \cite{spatialvla}; decoder contrasts remain within each backbone.

Cross-category/image stress pairs have high paired true-RGB token cosine. Validation selects the strictest threshold retaining at least 20 greedy object-disjoint pairs: $0.75$, giving 27 test pairs over 49 images/38 sessions. Nested pairs are fold-local (Supplement S4 fallback); neither applies a spectral-angle filter. Principal inference uses session-clustered pooled out-of-fold category bAcc, conditional on fitted models; image intervals test sensitivity. Three-seed spread describes optimization, not sampling uncertainty. A 38-session-node product-multiplier bootstrap handles two-session pair membership. The supplement supplies grids, seeds and class supports.

\section{Results}
Table~\ref{tab:session-common} reports recognition across 86 held-out sessions with 12 selection/60 refit epochs. The five-group held-session comparisons use validation-CE checkpoints: Table~\ref{tab:spectral-shape} tests spectral content under movement allowance 0.2, whereas Table~\ref{tab:same-residual-decoder}c,d test composition with unrestricted utility-selected scales. Table~\ref{tab:common-budget-main} tests calibration; Figure~\ref{fig:interventions} reports epoch-60 sensitivity. Reference-fixed readouts measure reference reuse; refitted readouts measure accessible information.

\subsection{Session-Held-Out Category Transfer}
Across held-out sessions (Table~\ref{tab:session-common}), base-fine improves pooled out-of-fold category bAcc by $0.071$ (95\% CI $[0.050,0.107]$) and proxy-state bAcc by $0.056$ ($[0.033,0.097]$). Direct replacement reaches cosine $0.991$ but category bAcc $0.814$, near reference $0.812$; base-fine reaches $0.883$, exceeding direct by $0.069$ ($[0.047,0.107]$). High reference similarity alone therefore misses the spectral gain. Fold-local selection/refitting precedes each session's single evaluation. The base contrast is primary; learned-method contrasts are exploratory, conditional on fitted predictions.

\begin{table*}[t]
\centering\small
\caption{SPECTRA-VLA improves recognition across 86 held-out acquisitions. Leave-one-session-out evaluation uses fold-local 12/60-epoch selection/refitting and one prediction per held-out object. Category/pooled-token proxy bAcc read pooled states; proxy labels are category-derived. Cos. measures false-color-reference similarity (all $\uparrow$). Differences are row minus SPECTRA-VLA, with conditional paired session 95\% intervals. High-anchor direct requires validation cosine $\geq0.94$. Gray identifies SPECTRA-VLA; bold marks favorable means, not significance.}
\label{tab:session-common}
\input{tables/tab_common_session_confirmation}
\end{table*}

The bridge combines reference anchoring with additional spectral discrimination. Projection attributes the residual to the pooled base's complement; norm control and validation-selected scale regulate executed movement. Removing projection improves recognition in some Table~\ref{tab:session-common} cells. Under a common movement allowance, base-fine instead achieves similar recognition with smaller disturbance tails (Table~\ref{tab:common-budget-main}); equal-achieved-movement comparisons separate this operating-point benefit from projection alone.

Recall improves in five of six categories; filament falls from $0.859$ to $0.848$ on 92 objects from 33 sessions. Every single-session deletion retains a positive category gain ($0.068$--$0.078$). Omitting pair utility changes 11 of 86 selections; pooling those 11 refits with the other 75 folds gives category/proxy bAcc $0.8815/0.9174$ versus $0.8829/0.9161$ over all 86 sessions, with both paired session intervals including zero. Training pair weights stay fixed. The supplement gives class counts and selection sensitivities.

Session-held-out band exclusion isolates information beyond the rendering: full spectra give category bAcc $0.886$ versus false-color-triplet $0.835$ (paired CI $[0.032,0.075]$); excluding those base bands gives $0.878$.

\subsection{Spectral Content Before and After Decoding}

Continuous spectra retain variation that category probabilities compress (Table~\ref{tab:spectral-shape}). Five outer groups test all 86 acquisitions once; inner session-disjoint classifiers supply training posteriors. Padding six-class codes to 136 coordinates matches adapter and selection settings, with 26,182 additional classifier parameters. The image split also tests one-hot codes; Section~6.4 compares recognition and disturbance.

\begin{figure*}[t]
\centering
\includegraphics[width=\textwidth]{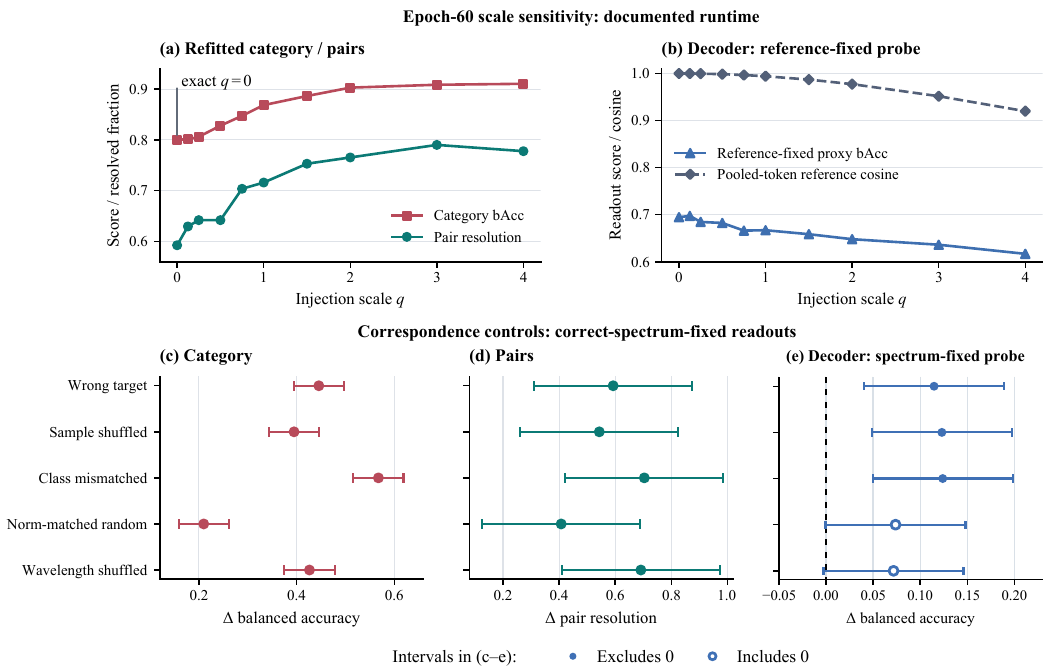}
\caption{Injection scale and correspondence affect recognition and decoder-readout reuse differently. (a,b) Epoch-60, image-disjoint three-seed curves use refitted category/pair and reference-fixed decoder probes. Table~\ref{tab:same-residual-decoder} changes checkpoint/scale, retaining the reference-trained probe. (c--e) Correct-minus-control differences retain correct-spectrum-trained readouts (``spectrum-fixed'' in e) and $q=(2,4,4)$ for seeds 0--2. Both (b,e) use recomputed decoder outputs; correct-spectrum bAcc in (e) is $0.710$. Filled/open markers denote intervals excluding/including zero. Session 95\% intervals are simultaneous across five controls per endpoint; pairs use 38-session-node product multipliers.}
\label{fig:interventions}
\end{figure*}

\textbf{Before decoding.} A linear readout of frozen category-trained tokens recovers each object's 120-band mean spectrum. Let $\bar x_i=x_i/\|x_i\|_2$ and $c_i$ be its category. Training-only means and scales define $\widetilde C_i=(C_{q,i}-\mu_{c_i}^C)/\sigma^C$ and $\widetilde x_i=(\bar x_i-\mu_{c_i}^x)/\sigma^x$. We measure
\begin{equation}
R^2=1-\frac{\sum_i\|\widehat{\widetilde x}_i-\widetilde x_i\|_2^2}
{\sum_i\|\widetilde x_i\|_2^2}.
\label{eq:within-category-recovery}
\end{equation}
Sums use test objects; the denominator is the training-category mean-spectrum baseline. All methods receive true-category conditioning and population scales of category-centered training residuals. Reconstruction targets train only the readout. Identical 4096-to-120 ridges use validation-MSE selection at allowance $0.2$ (Supplement S2.2).

\begin{table}[!htbp]
\centering\small
\caption{Continuous tokens retain more object-specific spectral variation than category codes. All blocks use validation-CE checkpoints and allowance $0.2$. Recovery $R^2$ and angle (radians) use train-fitted category conditioning (Eq.~\eqref{eq:within-category-recovery}); category bAcc uses the unconditioned category readout (Section~5.2). Held-session results pool five groups over 86 acquisitions; the final mapper comparison uses 16 sessions. Paired intervals resample acquisitions. Learned rows average three seeds; bAcc is shown where jointly evaluated. Bold favors means within blocks; gray identifies SPECTRA-VLA.}
\label{tab:spectral-shape}
\input{tables/tab_spectral_shape_main}
\end{table}

Table~\ref{tab:spectral-shape} separates continuous spectral content from category recognition. The recovery gain persists with signed pooling and an expanded ridge grid; true-category one-hot and shuffled training targets give $R^2\simeq0$ and $-0.184$.

Posteriors favor grouped category recognition ($0.882$ versus $0.861$); continuous states retain more spectral detail, with $\Delta R^2=0.465\ [0.429,0.502]$.

\textbf{Learned versus fixed-map residual.} In the separate 16-session mapper comparison, learning makes the retained spectrum more useful for recognition: category bAcc improves by $0.092$ (session CI $[0.017,0.174]$), while mean decoder KL decreases by $0.014$ ($[-0.021,-0.009]$). Projection, rescaling, own-token insertion and selection are identical (Table~\ref{tab:spectral-shape}; Supplement S2.4). The label-free 136-to-4096 Raw+NMF isometry retains higher $R^2$, separating spectral recoverability from learned discrimination. Reference-fixed/refitted decoder-logit probe bAcc is $0.634/0.694$ for learned and $0.646/0.595$ for fixed; both exploratory intervals include zero. This fixed residual omits the replacement control's pre-projection row LayerNorm.

\textbf{After frozen decoding.} Within-category spectral variation remains accessible in frozen decoder logits. Continuous/posterior inputs recover $R^2=0.096/0.085$ versus reference $0.083$, with continuous-minus-posterior gain $0.0113$ (66-session CI $[0.0042,0.0189]$). These jointly recomputed comparisons use validation-CE checkpoints, allowance $0.2$ and Eq.~\eqref{eq:within-category-recovery}'s conditioning. The 256-D pooled-token sensitivity gives $R^2=0.609$ before decoding (Supplement S2.2); width also contributes to cross-block differences. Each comparison matches readout width and conditioning.

\textbf{Target correspondence.} Correct target spectra support both waste-object category recognition and object-specific spectral recovery. Pooled-token category recognition and pair resolution exceed all five corruption controls (Figure~\ref{fig:interventions}c,d), with proxy-state gains in Supplement S2.1. Beyond category identity, the bridge preserves the target's own spectral variation, as shown by comparisons with same-category donors (Table~\ref{tab:same-residual-decoder}d). Epoch-60 interventions keep residual norms and correct-spectrum-trained readouts fixed. Joint decoder recomputation gives bAcc $0.710$ and gains $0.115$--$0.124$ over wrong-target, sample-permuted and class-mismatched inputs, with simultaneous lower bounds $0.040$--$0.050$. Random/band-permuted means are lower, with intervals crossing zero (Supplement S2).

\subsection{Composition at a Frozen Interface}
\textbf{Preserving the receiving grid.} At similar category recognition, own-token addition better supports reuse of the reference-trained decoder-logit probe. With spectral residuals and injection scale fixed, its balanced accuracy is $0.701$, compared with $0.590$ for pooled replacement on the image-disjoint test (Table~\ref{tab:same-residual-decoder}). Replacement moves the pool less yet has higher decoder-KL P95, $5.593$ versus $1.158$. Pooled proximity therefore misses changes in the token structure consumed downstream. These exploratory contrasts fix validation-CE checkpoints, residuals and $q=4$ (Eq.~\eqref{eq:replacement-grid}).

Zero-injection contrasts identify receiving-grid preservation; the image-disjoint reference-fixed four-state composition-by-injection interval includes zero (Table~\ref{tab:same-residual-decoder}b). Category-matched swaps below separately identify the content-dependent effect after decoding.

Refitted decoder-logit probes at these validation-CE checkpoints and unrestricted, utility-selected $q=4$ also improve decoder-logit probe bAcc over the reference from $0.695$ to $0.791$ (gain $0.096$, session CI $[0.031,0.201]$).

\begin{table*}[!t]
\centering\small
\caption{Own-token composition improves readout reuse and target-specific recovery. Arms share residuals/scales and recomputed logits. (a,b) Image-disjoint: 461 objects, CE-selected epochs $(3,6,4)$, $q=4$, reference bAcc $0.695$. (c,d) Five held-out groups: 2,291 objects/86 sessions, reference $0.681$, unrestricted validation-utility-selected scales. (d) Each cell separately train-fits centering, standardization and ridge; increment is own-minus-same-category-donor within-category $R^2$. Equal seed contributions share acquisition resampling, including donor nodes in (d); secondary intervals are unadjusted. Disp.: pooled movement; P95: per seed. Gray: base-fine; bold: favorable means, not significance. Supplement S2.4 gives influence.}
\label{tab:same-residual-decoder}
\textbf{(a) Identical spectral residual, different composition}\par\smallskip
\input{tables/tab_decoder_same_residual_main}
\par\smallskip Addition minus replacement: reference-fixed $+0.111\ [0.047,0.164]$; refit $+0.049\ [0.002,0.082]$.
\par\smallskip\textbf{(b) Composition-by-injection decoder-logit readouts}\par\smallskip
\input{tables/tab_composition_injection_main}
\par\smallskip Reference-fixed composition-by-injection interaction: $+0.013\ [-0.014,0.058]$.
\par\smallskip
\begin{minipage}[t]{0.53\textwidth}
\centering\textbf{(c) Held-session recognition and readouts}\par\smallskip
\input{tables/tab_session_composition_main}
\end{minipage}\hfill
\begin{minipage}[t]{0.45\textwidth}
\centering\textbf{(d) Decoder spectral recovery}\par\smallskip
\input{tables/tab_content_composition_recovery_main}
\end{minipage}
\par\smallskip Spectral content-by-composition interaction: $+0.0171\ [0.0121,0.0228]$.
\end{table*}

Across five groups, addition improves decoder-probe bAcc at similar category recognition (Table~\ref{tab:same-residual-decoder}c). The refitted gain concentrates in cardboard: 60 objects/24 acquisitions. Removing the first group's six cardboard objects retains gain $0.028\ [0.007,0.055]$; every session deletion stays positive (Supplement S2.4).

\textbf{Target-content--composition interaction.} Fixed adapters/scales and recipient projection/norm give own-minus-same-category-donor increments $0.0336/0.0165$ for addition/replacement, with interaction $0.0171\ [0.0121,0.0228]$ across 86 acquisitions (Table~\ref{tab:same-residual-decoder}d). Both shared recipient/donor resampling and the same-acquisition-only subset support this approximately doubled target-specific increment, distinct from total recovery or four-state classification.

\subsection{Calibrating Recognition and Disturbance}
Calibration compares recognition under a shared validation mean-displacement ceiling of $0.2$ (Table~\ref{tab:common-budget-main}). Utility selects scale at validation-CE checkpoints. Reference and altered logits are jointly recomputed with the common category ridge grid; test movement and divergence are measured afterward.

\begin{table*}[!t]
\centering\small
\caption{Recognition and disturbance at validation-CE checkpoints, using 461 objects from 66 sessions and the same category ridge grid. (a) Allowance $0.2$ gives achieved validation means $0.1438/0.1620$. Matching achieved mean target $0.15$ instead gives base-fine/no-projection category bAcc $0.871/0.871$ and KL P95 $0.163/0.160$ (Supplement S3.3). (b) Both inputs use the same composition and allowance $0.2$. Disp. and decoder KL are per-seed test 95th percentiles; $R^2$ reads within-category spectral shape. Gray identifies continuous SPECTRA-VLA. Bold marks favorable rounded means, not significance.}
\label{tab:common-budget-main}
\input{tables/tab_common_budget_main}
\end{table*}

Base-fine attains the same rounded category bAcc as the unprojected ablation, $0.880$, with smaller displacement and decoder-KL tails: P95 $0.188/0.152$ versus $0.213/0.182$ (Table~\ref{tab:common-budget-main}a). The common allowance $0.2$ yields validation mean displacements $0.1438/0.1620$. When achieved mean movement itself is matched, recognition and KL tails are similar (Supplement S3). Thus projection defines the base-complement residual, while calibration determines the measured disturbance of the selected operating point.

Under acquisition shift, validation mean-displacement target $0.15$ gives base-fine category bAcc $0.865$, test mean $0.150$, displacement P95 $0.194$ and decoder-KL P95 $0.164$ on 16 disjoint sessions. Validation sets the target; the supplement reports achieved movement, scales and tolerance.

\textbf{Continuous content versus category compression.} Category posteriors and continuous spectra favor different objectives. Under unrestricted utility selection, category and pooled-token proxy differences among continuous, hard-code and posterior inputs are unresolved. At the common allowance (Table~\ref{tab:common-budget-main}b), posterior-minus-continuous differences are $0.020$ ($[0.001,0.035]$) in category bAcc and $-0.065$ ($[-0.086,-0.047]$) in decoder-KL P95. Decoder-logit probe differences remain unresolved. Posteriors favor classification and smaller disturbance, whereas continuous input retains within-category spectral variation beyond the six-category decision. Selecting scale for validation recovery $R^2=0.8$ yields test recovery $0.801$, category bAcc $0.871$ and decoder-KL P95 $0.087$ (Supplement S2.1--S2.3).

\subsection{Checkpoint and Readout Sensitivity}
\label{sec:checkpoint-sensitivity}
At epoch 60, validation-selected $q=(2,4,4)$ improves category/pair scores to $0.905/0.778$ from $0.801/0.593$, while reference-fixed bAcc falls $0.695\to0.617$ ($-0.077\ [-0.118,-0.035]$). The pair-gain session interval $[-0.030,0.443]$ includes zero (Figure~\ref{fig:interventions}a,b). Earlier CE checkpoints with $q=4$ give $0.701$ in the same runtime/readout. Supplement S3 gives cached sensitivities; S6 reports short-budget held-session scores $0.684/0.705$ for altered/reference states.

\begin{figure*}[t]
\centering
\includegraphics[width=\textwidth]{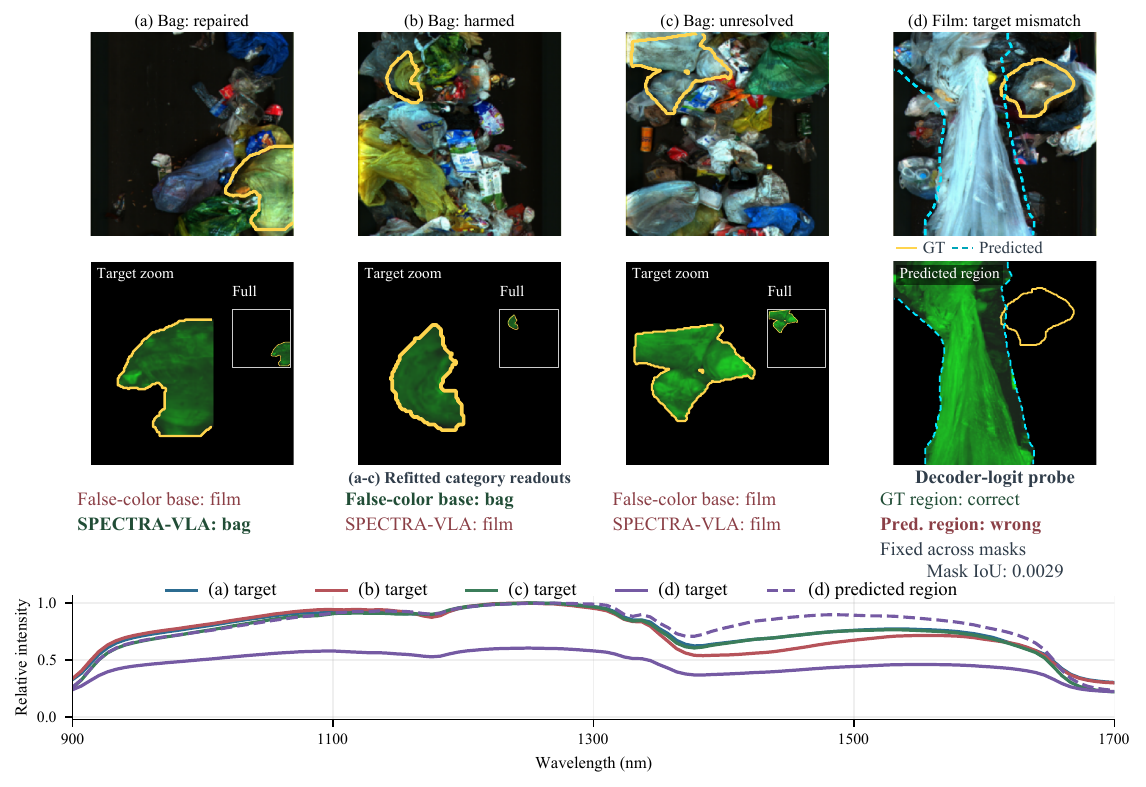}
\caption{Category repair, injection-induced harm, unresolved recognition and target-region failure on fixed-test examples. True RGB marks targets. Lower views (a--c) enlarge targets for display only; side views show original full-frame inputs at unchanged intensities. Refit category predictions are bold when correct. In (d), ground-truth (solid gold) and predicted (dashed cyan) regions have IoU $0.0029$; the ground-truth-region-trained decoder-logit probe changes from correct to wrong. Spectra (a--c) use each target's peak; both (d) curves share the larger region peak. Wavelengths approximate an uncalibrated uniform grid. Outcome-selected cases do not estimate prevalence.}
\label{fig:qualitative-pair}
\end{figure*}

\subsection{Backbone, Grounding, and Failure Analysis}
On the fixed image-disjoint split, SpatialVLA reproduces the representation pattern: category/proxy balanced accuracy rises from $0.812/0.871$ to $0.914/0.947$ at cosine $0.985$. Paired session intervals for the gains are $[0.043,0.177]$ and $[0.025,0.127]$, respectively. Reference-fixed decoder-logit probe balanced accuracy remains lower than the reference ($0.673$ versus $0.718$). The supplement reports direct-retargeting and constraint comparisons, together with spectral interventions on its distinct action vocabulary.

Target quality determines which spectrum enters the bridge. Four-state decoder-logit probes are trained separately on each representation's ground-truth-region logits and kept fixed across mask perturbations. One-pixel erosion gives base-fine/base bAcc $0.732/0.683$, while a four-pixel shift gives $0.638/0.643$. Poor class-oracle predicted masks, with mean IoU $0.218$, lower base-fine below the base ($0.362/0.468$), even with the category supplied. Each condition re-encodes its region with unchanged injection scales.

Base-fine retains $0.975$ of base-correct category predictions and repairs $0.542$ of base-wrong predictions. Figure~\ref{fig:qualitative-pair}(a--c) shows refitted category predictions. Separate reference-fixed decoder-logit probe predictions are correct for the base and base-fine in (a,b), but only for the base in (c). In (d), predicted-mask IoU $0.0029$ changes the region-trained decoder-logit probe prediction from correct to wrong. Category correction and reuse of a reference-trained logit readout differ even on the same object.

Published spectral encoders provide a complementary recognition comparison under the same ridge grid; the supplement reports five spectral encoders and a raw-spectrum reference, rendering sensitivity and external datasets. Partially loaded DOFA/HyperSIGMA are excluded from peer ranking. Mendeley Wood/SpecTex use expanded ridge/PCA readouts of learned residual features to test dataset-specific augmentation, separately from native-width VLA composition.

\section{Discussion}
Spectral enrichment and interface preservation are complementary. Continuous inputs retain variation lost by category compression (Table~\ref{tab:spectral-shape}); receiving-token controls tie decoder stability to grid structure rather than pooled proximity (Table~\ref{tab:same-residual-decoder}). Across all five acquisition groups, same-category swaps link receiving-token composition to a larger recovery increment from the target's own spectrum, separately from the no-injection grid benefit.

Learning supplies discrimination; own-token addition retains the receiving grid. Projection specifies a base-complement direction, while norm control and validation calibration regulate executed movement. Common-allowance and equal-movement comparisons attribute smaller disturbance tails to the selected operating point (Table~\ref{tab:common-budget-main}; Supplement S3). Posterior codes favor classification; continuous input retains within-category shape. Learned residuals improve recognition and mean decoder KL over fixed maps, with a recovery trade-off. These results separate content, composition and calibration choices.

HyA-T also freezes both pretrained components~\cite{hyat}. SPECTRA-VLA preserves receiving tokens at native width, as shown by matched spectral-content, composition and calibration tests. Its auxiliary heads serve training only.

\section{Conclusion}
SPECTRA-VLA upgrades frozen VLA interfaces at native token width. Category bAcc rises $0.812\to0.883$ across 86 held-out acquisitions; a second backbone reproduces enrichment. Continuous input retains within-category spectral variation. Own-token composition improves fixed-readout reuse at similar recognition and approximately doubles target-specific recovery increments across five acquisition groups. The frozen encoder/decoder and exact zero-injection reference are preserved, coupling new spectral content to the pretrained receiving structure.

\textbf{Code and data availability.} Code is available at \url{https://github.com/jesse1029/SPECTRA-VLA}. Grouped predictions, donor identities and numerical results are available at \url{https://www.dropbox.com/scl/fi/xkp5xlb8ldjbmrgljx4ku/SPECTRA-VLA_numerical_data_only.zip?rlkey=ycpmxfvmaxzquax6xw0zhe8aa&dl=0}. Supplement S1 specifies checkpoints and inference.
\section*{Acknowledgment}
This study was supported in part by the National Science and Technology Council (NSTC), Taiwan, under grants 115-2627-M-006-001, 115-2218-E-A49-023, and 115-2634-F-A49-014. The author thanks the National Center for High-performance Computing (NCHC) of National Applied Research Laboratories (NARLabs) in Taiwan for providing computational and storage resources.

OpenAI Codex and ChatGPT~\cite{openai_codex,openai_chatgpt} assisted with implementation, experimental analysis, and manuscript drafting and revision. Their use included implementation and analysis associated with the method and experiments, and revision of the main paper and supplementary material. The author remains responsible for the scientific claims, reported results, and final manuscript.

\bibliographystyle{IEEEtran}
\bibliography{spectra_refs}
\end{document}


\maketitle

\input{supplement_compact_body}

\twocolumn[]
\raggedbottom
\IEEEtriggeratref{11}
\bibliographystyle{IEEEtran}
\bibliography{spectra_refs}

%% file: tables/tab_session_composition_main.tex
\begin{tabular}{@{}lrrrr@{}}
\toprule
\textbf{Readout} & \textbf{Add} & \textbf{Replace} & $\boldsymbol{\Delta}$ & \textbf{95\% CI} \\
\midrule
Category & 0.876 & \textbf{0.877} & -0.001 & $[-0.0068,0.0035]$ \\
Ref.-fixed & \textbf{0.683} & 0.658 & +0.026 & $[0.0004,0.0572]$ \\
Refit & \textbf{0.735} & 0.700 & +0.035 & $[0.0148,0.0656]$ \\
\bottomrule
\end{tabular}

%% file: tables/tab_content_composition_recovery_main.tex
\begin{tabular}{@{}lrrr@{}}
\toprule
\textbf{Composition} & \textbf{Correct} & \textbf{Donor} & \textbf{Increment $\uparrow$} \\
\midrule
\rowcolor{proposedgray} \textbf{Base-fine} & 0.1010 & 0.0674 & \textbf{0.0336} \\
Replacement & 0.0966 & 0.0801 & 0.0165 \\
\bottomrule
\end{tabular}

%% file: supplement_compact_body.tex
\section{Experimental Details}
\label{sec:supp-details}
\noindent\textit{AI assistance.} OpenAI Codex and ChatGPT~\cite{openai_codex,openai_chatgpt} assisted with implementation, analysis, and preparation of this supplement; the main paper's Acknowledgment describes their use and the author's responsibility.

SPECTRA-VLA enriches the frozen VLA visual-token interface with target-corresponding spectral information. The following experiments distinguish spectral content, receiving-token composition and injection calibration. Category and category-derived proxy readouts classify pooled states; decoder-logit probes classify frozen first-position action-symbol logits. Gray shading identifies SPECTRA-VLA, and bold marks favorable displayed means within comparable groups, including ties. Paired intervals assess statistical support separately.

\subsection{Observations, Targets and Partitions}
SpectralWaste \cite{spectralwaste} contains 514/167/171 training/validation/test source images with synchronized RGB, a $224\times256\times256$ band-first HSI cube and category masks. Cubes retain their original intensity scale in float32. Linear interpolation maps 224 bands to 120, sampling source position $(k+\tfrac12)224/120-\tfrac12$ for output band $k$. HSI and RGB are bilinearly resized from $256^2$ to $224^2$ with pixel-center interpolation; masks use nearest neighbors. Category-specific four-connected components of at least 100 resized pixels define targets. This gives 1,392/442/461 objects in 479/156/159 images; 35/11/12 source images contain no retained component.

Images are zeroed outside the target, without cropping, padding or additional registration. The fixed false-color view uses bands $[119,60,0]$, divides each channel by $65535+10^{-6}$, clips to $[0,1]$ and casts $255$ times the result to uint8. This spans the sensor's index range; it is distinct from paired visible RGB. Area pooling reduces the target mask to the $16\times16$ token grid. Raw occupancy $\widetilde m_{it}$ and injection weight $m_{it}=\widetilde m_{it}/(\max_t\widetilde m_{it}+10^{-6})$ are kept distinct.

The six labels are operational waste-object categories. Table~\ref{tab:supp-material-state-map} defines their four-state reduction; only state 1 merges categories. State-0/1/2/3 counts are 35/760/200/397, 15/256/54/117 and 10/254/67/130 in training, validation and test. Predicting a category and applying this mapping provides the category-lookup diagnostic.
\begin{table}[!tbp]
\centering\small
\caption{Waste-object categories, object counts and the fixed four-state proxy mapping. Category IDs and names follow SpectralWaste; the grouping is defined by the evaluation.}
\label{tab:supp-material-state-map}
\input{tables/tab_material_state_mapping}
\end{table}

An object, not an individual wavelength, is the prediction unit. The image-disjoint split contains 2,295 objects in 794 images; its train/validation/test partitions occupy 82/63/66 overlapping acquisition sessions. Sessions follow the source-image acquisition prefix. Session-held-out evaluation instead uses 2,291 objects in 792 images from the 86 sessions represented in fixed training or test. It includes 438 fixed-validation objects from overlapping sessions and excludes four objects from two validation-only sessions. Each outer fold holds out a complete acquisition. Thus image holdout separates images, whereas session holdout separates all images acquired together.

\subsection{Training-Fitted Spectral Features}
\label{sec:supp-spectral-features}
The 120-D raw spectrum averages target pixels. The sensor range is 900--1700~nm, but the release provides no per-band calibration vector; wavelength-dependent analyses use an approximate uniform grid. Inputs always follow the source-index interpolation above.

The main 136-D input joins 120 raw bands and 16 NMF abundances. Each training partition fits per-band standardization and nonnegative 99th-percentile scaling. NMF uses NNDSVDa initialization, coordinate-descent Frobenius updates, at most 2,000 iterations, tolerance $10^{-4}$ and seed $20260710+s$. Abundances are divided by their sum plus $10^{-8}$; raw and abundance blocks are standardized separately. All fitted transforms remain within the relevant training partition and are refitted within each outer-training population.

\subsection{Adapter, Loss and Token Composition}
The rank-$r$ mapper is
\[
\phi_r(s)=W_{\rm up}\operatorname{GELU}
  \bigl(W_{\rm down}\mathrm{LN}_{\rm aff}(s)\bigr),
\]
with affine input LayerNorm, a biased down projection and bias-free up projection. The main model uses $r=64$, output width $d=4096$ and norm budget $\rho=0.1$: 271,184 mapper parameters and 320,348 including auxiliary heads. Non-affine composition LayerNorm uses $\epsilon=10^{-5}$; projection and norm control use $\delta=10^{-6}$.

Writing $B^\circ=\mathrm{LN}_\epsilon(B)$, training uses $T_q=\mathrm{LN}_\epsilon(B^\circ+qD)$ for $q>0$ and $T_0=B^\circ$. The composed and base losses share linear head $h_c$; the fine head $h_f$ is separate:
\[
\begin{aligned}
\mathcal L={}&\mathrm{CE}_w(h_c(T_q),y)
 +\lambda_f\mathrm{CE}_w(h_f(D),y)\\
 &+0.1\,\mathrm{CE}_w(h_c(T_0),y)
 +10^{-4}\operatorname{mean}(D^2).
\end{aligned}
\]
The fixed-split default is $\lambda_f=1$. For $C$ training classes with counts $n_c$, mean-one class weights are $w_c=C/(n_c\sum_k n_k^{-1})$. Each minibatch reduces classification losses as $\sum_i\omega_i[-w_{y_i}\log p_{i,y_i}]/\sum_i\omega_i$, with $\omega_i=4$ on training stress-pair endpoints and 1 elsewhere. The denominator excludes class weights; the residual penalty is unweighted. Each item draws $q\sim\mathcal U(0,4)$. The $T_0$ loss updates $h_c$ but has no residual-path gradient. Auxiliary heads are discarded after training.

Evaluation supplies the full 256-token grid to the frozen decoder:
\[
A_{it}(D,q)=
\begin{cases}
G_{it},&q=0,\\
(1-m_{it})G_{it}
 +m_{it}\mathrm{LN}_\epsilon(G_{it}+qD),&q>0.
\end{cases}
\]
Pooled readouts use $C_q=\mathrm{LN}_\epsilon(\operatorname{pool}_m A(D,q))$. This executed pool differs from the training surrogate $T_q$. Replacement uses $R_{it}(v)=(1-m_{it})G_{it}+m_{it}\mathrm{LN}_\epsilon(v)$, broadcasting one vector over target tokens. The no-LayerNorm ablation substitutes $G_{it}+qD$ for the normalized branch. All operators preserve zero-weight tokens.

\subsection{Optimization and Fixed-Split Selection}
All matched learned representations use seeds $\{0,1,2\}$, batch size 256, seeded epoch permutations and a 60-epoch budget. AdamW uses learning rate $2\times10^{-3}$, weight decay $10^{-3}$, $\beta=(0.9,0.999)$ and $\epsilon_{\rm Adam}=10^{-8}$, without a scheduler; gradient norm is clipped at 1. Linear layers use seeded PyTorch default initialization and affine LayerNorm uses unit gain/zero bias. Fixed-epoch comparisons retain epoch 60. Mechanism comparisons select the minimum class-weighted composed-head validation CE at $q=1$, excluding auxiliary losses, before scale selection.

The injection grid is
\[
\mathcal Q=\{0,0.125,0.25,0.5,0.75,1,1.5,2,3,4\}.
\]
For each seed and candidate, category ridge regularization is first chosen by validation bAcc from $\{0.01,0.1,1,10,100\}$, preferring the smaller coefficient on exact ties. Candidates are ordered by $(\mathrm{bAcc}+0.25\,\mathrm{Pair},\mathrm{bAcc},-q)$; a missing pair value contributes zero. Proxy and decoder ridge coefficients are selected independently after fixing the representation. Fixed replacement rows use scale one. Random-label adapters inherit the correctly supervised model's scale. The final-epoch base-fine choices are $q=(2,4,4)$; category/proxy/decoder ridge coefficients are $(100,100,100)/(100,100,10)/(100,100,100)$.

Spectrum-only retargeting uses $\alpha\in\{0,0.01,0.05,0.1,0.25,0.5,1,2,5,10\}$ and the analogous tuple with $-\alpha$, selecting $(0.5,10,0.5)$. Its category/proxy/decoder ridge choices are $(10,10,100)/(100,100,100)/(10,100,100)$. Base-conditioned replacement adds $\{25,50,100\}$ to this alignment grid, requires validation cosine $\geq0.94$, then maximizes category-plus-pair utility, cosine and smaller $\alpha$ in order. It selects $(0.01,0.05,0.05)$; category and proxy ridges are 100 for all seeds. Its decoder diagnostic uses the common reference-fixed ridge 10.

\subsection{Pooled-State and Decoder Readouts}
For $X\in\mathbb R^{256\times16}$ obtained by reshaping a 4096-D state, signed pooling is
\[
P_k(X)=16^{-1/2}\sum_{j=1}^{16}S_{kj}X_{kj},
\]
where $S$ is a fixed Rademacher sign pattern with seed 20260712. Other widths use the corresponding fixed projection $xR/\sqrt{256}$. Training-fitted standardization precedes class-balanced ridge with an iterative least-squares solver and the five-value validation grid above. OpenVLA's 256 logit coordinates are retained up to signs; SpatialVLA's 8,194 coordinates are projected to width 256. Prototype prediction uses the same pooled, standardized features $u(x)$:
\[
\mu_c=n_c^{-1}\sum_{i:y_i=c}u(x_i),\qquad
\hat y=\arg\max_c \nu(u(x))^\top\nu(\mu_c),
\]
with $\nu(v)=v/(\|v\|+10^{-8})$.

Decoder-logit probes are external four-state classifiers on frozen logits. A \emph{refitted} probe learns separately for each representation. A \emph{reference-fixed} probe retains the reference-trained standardizer and classifier. Correspondence controls retain the \emph{correct-spectrum-trained} probe. These measure accessible information, reference decision-function reuse and content dependence, respectively.

Official weights and matching tokenizers/processors use revisions \href{https://huggingface.co/openvla/openvla-7b}{\texttt{47a0ec7fc4ec}} (OpenVLA) and \href{https://huggingface.co/IPEC-COMMUNITY/spatialvla-4b-224-pt}{\texttt{886c4557dfde}} (SpatialVLA). OpenVLA-7B \cite{openvla} receives \texttt{In: What action should the robot take to sort the object into the correct bin?\textbackslash nOut:} followed by space token 29871. Frozen features pass through its pretrained visual-to-language projector. The 256 composed embeddings and attention-mask entries follow the beginning-of-sequence embedding; first-position logits use the last real prompt token. The unpadded action reference is 32,000 versus 32,064 runtime logits; $j_b=31{,}999-b$, $b=0,\ldots,255$, gives IDs 31,744--31,999. SpatialVLA-4B-224 \cite{spatialvla} uses the same instruction without \texttt{In:/Out:}, its Gemma tokenizer/processor, 256 image positions and 8,194 action coordinates starting at 257153. The probe reads all coordinates, versus native three-token decoding.

KL uses temperature-one softmax renormalized over the specified action support. Token cosine averages samplewise target-pooled cosines; displacement averages $\|z-z^{\rm ref}\|/(\|z^{\rm ref}\|+10^{-12})$. Backbone-specific KL supports differ, so their magnitudes are interpreted within each backbone.

\subsection{Numerical Settings}
Optimization and composition normalization use float32; frozen inference/grid storage use bfloat16/float16. Joint OpenVLA uses PyTorch 2.6.0, CUDA 12.4, Transformers 4.40.1, SDPA, batch four, evaluation mode and no key--value cache (reference bAcc $0.694685$). Earlier cached sensitivities match precision/batch/cache settings, requesting SDPA with eager fallback (reference bAcc $0.704116$). Original versions are incompletely documented; retained logits/predictions, splits, fitted readouts and table scripts support reproduction. Eight-observation maximum/mean absolute logit differences are $0.40625/0.077215$; top-1 symbols agree. Each contrast uses one configuration and reference. SpatialVLA uses Transformers 4.47.0.

\section{Spectral Content}
\label{sec:supp-interventions}
\subsection{Correspondence and Category-Code Controls}
Correspondence controls preserve residual width and norm while changing spectral content. Category, proxy and decoder heads are fitted/selected with correct spectra and remain fixed. Epoch-60 adapters use $q=(2,4,4)$. The separately trained no-fine-loss control inherits these scales here; its independently selected/refitted comparison uses $(4,4,4)$.

Wrong-target donors come from the same image, else the same session, else another category, with target index modulo donor count. Class mismatch cycles to the next sorted category. Partitionwise sample derangements use seed $20260710+s+1009p$ for partition $p$, at most 64 draws, then a one-position rotation. Norm-matched Gaussian residuals use seed $99173+s$; band permutation uses $773+s$. Foreign/random residuals are reprojected against the recipient base and matched to its correct-residual norm.

\begin{table*}[!tp]
\centering\small
\caption{Correct spectral correspondence supports recognition and decoder information. (a) Correct-spectrum-fixed readouts at epoch 60. (b) Simultaneous paired intervals over five cached-logit controls: shared image/session clusters, both endpoint-session memberships for pairs. (c) Joint recomputation with unchanged correct-spectrum heads; session intervals are primary, image intervals sensitivity. SD measures optimization spread.}
\label{tab:supp-interventions}\label{tab:supp-intervention-clustered}\label{tab:supp-content-runtime}
\textbf{(a) Recognition and cached-logit decoder sensitivity}\par\smallskip
\input{tables/tab_spectral_interventions}
\par\medskip\textbf{(b) Correct-minus-control simultaneous intervals}\par\smallskip
\input{tables/tab_intervention_clustered}
\par\medskip\textbf{(c) Jointly recomputed decoder contrasts}\par\smallskip
\input{tables/tab_content_runtime_confirmation}
\end{table*}

Correct spectra outperform wrong-target, sample-permuted and class-mismatched inputs in the jointly recomputed decoder comparison. Norm-random and band-permuted means are lower, but their simultaneous intervals cross zero. Intervals use 100,000 shared $\mathrm{Exp}(1)$ cluster draws (seed 20261005), seed averaging within draws and the maximum centered deviation across five contrasts. Pair inference uses the product of distinct endpoint-session multipliers and a single multiplier within one session (Section~\ref{sec:supp-pairs}).

A same-category/different-session donor control keeps training/validation states unchanged and replaces only test residuals. An additional linear predicted-label control fits train-only raw-plus-NMF ridge ($\alpha=1$); training codes are in-sample and validation/test codes out-of-sample. A learned $6\times64$ embedding and bias-free $64\times4096$ projection produce a base-orthogonal residual without norm rescaling, with selected $q=(0.75,0.75,0.5)$. Base-fine/donor/predicted-label category/proxy bAcc is $0.904645/0.911276$, $0.886397/0.915321$ and $0.860065/0.877341$ under ridge 1. Base-fine-minus-donor intervals span zero; differences from predicted labels are $0.044580$ [$0.017565,0.081454$] and $0.033935$ [$0.006015,0.061737$]. Donors share the base-fine head, while predicted-label states refit it.

Stronger category-code controls use a 136--128--64--6 ReLU classifier, inverse-frequency CE, AdamW with learning rate $10^{-3}$/weight decay $10^{-4}$ and 100 epochs. Five session-grouped folds supply out-of-fold training codes; full-training refitting supplies validation/test codes. Six-dimensional one-hot or posterior codes are padded to 136 and passed through the same rank-64, 320,348-parameter bridge, projection, norm bound, CE-checkpoint rule, scale grid and readout. The upstream classifier adds 26,182 parameters. Hard/posterior/continuous scales are $(1.5,0.75,1)/(1,4,1)/(4,4,4)$. Category/proxy differences are small, with all continuous-versus-code paired intervals spanning zero.
\begin{table*}[!tp]
\centering\small
\caption{Continuous spectra and matched category codes offer distinct operating points. (a) Utility-selected scales with refitted pooled readouts. (b) Common validation displacement allowance $0.2$, with jointly recomputed fixed/refitted decoder probes. The code classifier adds 26,182 parameters; paired decoder uncertainty appears in Table~\ref{tab:supp-composition-decoder-intervals}.}
\label{tab:supp-category-codes}\label{tab:supp-input-code-decoder}
\textbf{(a) Category-code scale comparison}\par\smallskip
\input{tables/tab_r6_categorical_codes}
\par\medskip\textbf{(b) Common movement allowance}\par\smallskip
\input{tables/tab_decoder_input_codes}
\end{table*}
At allowance $0.2$, posterior scales are $(1,1,1)$, mean validation displacement $0.1084$ and category bAcc $0.899275$. At allowance $0.1$, scales are $(0.75,0.75,0.75)$ and category bAcc $0.889462$. Posterior codes favor classification and smaller disturbance; continuous spectra favor spectral-shape retention, examined next.

\subsection{Within-Category Spectral-Shape Recovery}
\label{sec:supp-shape}
A linear probe recovers the 120-band object-average spectrum from fixed category-trained states. Shape targets are $\bar x_i=x_i/\|x_i\|$. For states or spectra $a_i$, subtract their training-category mean $\mu_c^a$, then divide by the coordinatewise population standard deviation of these category-centered training residuals. Floors are $10^{-6}$ for states/logits and $10^{-8}$ for spectra. True-category conditioning removes between-category variation from both predictors and targets.

A training-only ridge with an intercept predicts standardized targets from 4096-D executed pools at the existing allowance-$0.2$ scales. Validation MSE selects $\{10^{-4},10^{-3},10^{-2},10^{-1},1,10,100,1000\}$, preferring smaller coefficients on ties. With standardized spectral target $\widetilde x$,
\[
R^2=1-\frac{\sum_{\rm test}\|\widehat{\widetilde x}_i-\widetilde x_i\|^2}
                  {\sum_{\rm test}\|\widetilde x_i\|^2}.
\]
The denominator is the error of training-category mean spectra. Spectral angle, in radians, is measured after reversing standardization and adding the category mean. This measures within-category object-average shape.

\begin{table*}[!tp]
\centering\small
\caption{Continuous payloads retain spectral shape beyond category identity. (a) Fixed-state recovery and uniformly expanded ridge-grid sensitivity. RMSE uses standardized targets; angle is in radians. (b) Paired cluster intervals; the first session comparison is primary within this post-hoc analysis. (c) Recovery from the 256 frozen decoder logits at the same allowance-$0.2$ states. Seed SD is optimization spread.}
\label{tab:supp-spectral-shape-sensitivity}\label{tab:supp-spectral-shape-intervals}\label{tab:supp-decoder-spectral}
\textbf{(a) Pooled-state recovery and sensitivities}\par\smallskip
\input{tables/tab_spectral_shape_sensitivity}
\par\medskip\textbf{(b) Paired differences}\par\smallskip
\input{tables/tab_spectral_shape_intervals}
\par\medskip\textbf{(c) Recovery after frozen decoding}\par\smallskip
\input{tables/tab_decoder_spectral_recovery}
\end{table*}

Continuous-minus-posterior recovery is $0.450000$, with session interval $[0.397758,0.504428]$ and image interval $[0.398236,0.506449]$. Shared resamples recompute seed-specific error ratios, then average paired differences (10,000 draws, seed 20261005; 66 sessions or 159 images). Spectrum-input/category-only/shuffled-target controls give $R^2\simeq1,0,-0.184459$. Recovery persists under 256-D signed pooling. Adding $10^4,10^5,10^6$ uniformly to the ridge grid yields continuous/posterior $0.823875/0.359549$, paired difference $0.464326$ [$0.416840,0.514945$]. Frozen-logit recovery improves by $0.011339$ [$0.004193,0.018923$] under session resampling. These conditional probes leave adapters, scales and the VLA unchanged.

\subsection{Spectral-Recovery Calibration}
Validation recovery targets are $R^2=0.8$ (primary) and $0.7$ (secondary). Calibration takes the first adjacent scale-grid bracket crossing the target, then bisects for at most 18 iterations to absolute error $\leq0.002$, selecting spectral ridge by validation MSE at each scale. Primary scales are $(1.015625,1.031250,1.031250)$; secondary scales are $(0.4921875,0.515625,0.515625)$. The primary point reaches test $R^2=0.801$, category bAcc $0.871$, displacement P95 $0.146$ and decoder-KL P95 $0.087$; secondary test recovery is $0.713095$. Thus scale can be chosen for an information requirement rather than maximum recognition.

\subsection{Session-Disjoint Mechanism Tests}
Five acquisition groups (seed 20261006) hold out all 86 eligible sessions once: 16/16/18/18/18 sessions and 329/378/597/647/340 objects, totaling 2,291. Each outer-training set uses the first inner five-fold group split (seed 20261007); five session-disjoint classifiers supply training posteriors. All transforms and readouts fit training data. Both inputs share rank 64, 320,348 bridge/head parameters, three seeds, $\rho=0.1$ and 60 epochs with minimum-validation-CE checkpoint selection. Continuous/posterior epochs span 3--27/3--13, unrestricted scales 1--4/0.75--4 and allowance-$0.2$ scales 1--1.5/0.75--1.5. S1 defines category-plus-pair selection; recovery minimizes validation MSE. This schedule differs from the principal 86-fold category evaluation.

\begin{table*}[!tp]
\centering\small
\caption{Spectral content and composition across five held-out groups/86 sessions. (a) Allowance-$0.2$ recovery/category bAcc; angle is in radians. (b) Shared residuals; fold-local scales maximize unrestricted validation utility. Reference-fixed probes retain reference training; each refit probe trains on its cell's logits, including donors. Donors change correspondence, preserving category. Values pool out-of-fold objects after equal seed averaging.}
\label{tab:supp-session-spectral}\label{tab:supp-session-composition}
\textbf{(a) Continuous spectra versus category posteriors}\par\smallskip
\input{tables/tab_session_spectral_confirmation}
\par\medskip\textbf{(b) Composition with and without injection}\par\smallskip
\input{tables/tab_session_composition_confirmation}
\end{table*}

Continuous input preserves more within-category variation than posterior codes (paired $R^2$ gain $0.465$ [$0.429,0.502$]), while posterior codes favor classification by $0.021$ [$0.009,0.034$]. Table~\ref{tab:supp-session-spectral} gives all means and the false-color reference.

At similar category recognition, addition improves refitted decoder bAcc by $0.035$ [$0.015,0.066$] (Table~\ref{tab:supp-session-composition}). Cardboard accounts for the gain: 60 objects/24 acquisitions; the other three recalls slightly favor replacement (Table~\ref{tab:supp-session-influence}). Removing the first group's six cardboard objects retains refit gain $0.0283$ [$0.0067,0.0551$] and reference-fixed gain $0.0271$ [$-0.0003,0.0637$]. Every session deletion preserves refit gains ($0.0310$--$0.0414$). All 10,000 full-data bootstrap draws contain four states.
\begin{table*}[!tp]
\centering\small
\caption{Paired uncertainty and class contributions across 86 acquisitions. (a) Shared resampling, 10,000 draws, seed 20261012: node-product multipliers for donor contrasts, multinomial session bootstrap otherwise. Seeds contribute equally. Content-by-composition and reference-fixed addition/replacement are primary; other intervals are unadjusted. Spectral rows: within-category $R^2$; state rows: bAcc. (b) Refitted proxy-state recall/support. Bold marks higher displayed recall, including ties; contributions differ from sampling uncertainty.}
\label{tab:supp-session-mechanism-intervals}\label{tab:supp-session-influence}
\textbf{(a) Paired session differences}\par\smallskip
\input{tables/tab_session_mechanism_intervals}
\par\medskip\textbf{(b) Class support and recall}\par\smallskip
\input{tables/tab_session_influence}
\end{table*}

The separate fixed-map and achieved-movement diagnostics retain the first outer group (329 objects/16 sessions). A label-independent fixed map uses a $4096\times136$ Gaussian reduced-QR isometry (seed 20260710), followed by unchanged projection and norm control, without preprojection LayerNorm. Allowance $0.2$ selects fixed-map scales $(1.5,1.5,1.5)$ versus learned $(1,1.5,1)$. Mean-displacement calibration to validation target $0.15$ within $10^{-4}$ selects $(1.3809,1.3809,1.3809)$ and category ridge 100; test mean displacement/category bAcc/displacement P95/KL P95 is $0.150/0.865/0.194/0.164$.

\paragraph*{Content-specific composition effect}
Within each outer group, the same adapters and selected scales cross own-token addition or pooled replacement with correct or same-category donor spectra. Partition/category derangements use seed $202610080+s$, shuffled session order, stable image order and cyclic donor shifts, favoring within-session donors. Donors are reprojected against recipient bases and norm-matched; normalized-dot/relative-norm errors are below $10^{-5}$. Every cell independently fits training-category feature centering, standardization and a 256-to-120 ridge; common recipient spectral targets use the conditional recovery definition above.

For addition/replacement $(A/R)$ and correct/donor $(C/W)$, the primary interaction is
\[
I=(R^2_{A,C}-R^2_{A,W})-(R^2_{R,C}-R^2_{R,W}).
\]
Donors cross acquisitions for 11.6\% of objects. Shared node multipliers $\xi_s\sim\mathrm{Exp}(1)$ weight same-acquisition pairs by $\xi_s$ and distinct pairs by $\xi_s\xi_t$, with equal seed contributions to pooled error ratios. Inference conditions on fitted models and fixed donors. Own/donor recovery is $0.1010/0.0674$ for addition and $0.0966/0.0801$ for replacement: increments $0.0336/0.0165$ give $I=0.0171$ [$0.0121,0.0228$] (Table~\ref{tab:supp-session-mechanism-intervals}). Same-acquisition donors in all seeds retain 2,026 objects/76 sessions and $I=0.0162$ [$0.0122,0.0210$]; dyadic overlap-sandwich sensitivity gives $[0.0118,0.0224]$. Omitting the first group's six cardboard objects gives $I=0.0170$; recipient-session deletion range is $0.0161$--$0.0183$. Correct-content totals differ by $0.0044$ [$-0.0093,0.0188$], separating content-dependent increments from total recovery.

Same-category donors retain or improve proxy recognition because the labels remain unchanged (Table~\ref{tab:supp-session-composition}); recipient-spectrum recovery instead tests object correspondence. Two excluded validation-only acquisitions contain four state-1 objects, evaluated by each frozen fold fit as a separate sensitivity. Repeated predictions are not additional sampling units.

\section{Composition and Calibration}
\subsection{Alignment and Native-Width Composition}
Spectrum-only retargeting minimizes
\begin{equation}
\begin{aligned}
z_{\rm ret}&=\mathrm{LN}_\epsilon(\phi_r(s)),\\
\mathcal L_{\rm ret}
 &=\mathrm{CE}_w(h(z_{\rm ret}),y)
   +\alpha d^{-1}\|z_{\rm ret}-B^\circ\|^2.
\end{aligned}
\label{eq:supp-direct}
\end{equation}
Endpoint weights apply to both terms; class weights apply only to CE. Direct projection instead uses $1.1\,\mathrm{CE}_w(h_c(z),y)+\mathrm{CE}_w(h_f(\phi_r(s)),y)+10^{-4}\operatorname{mean}(\phi_r(s)^2)$. Fine-only normalizes the constrained residual as a replacement. Concatenation-projection uses the joint input $[B^\circ;s]$ at native output width; feature concatenation retains width $d+136$.

Base-conditioned replacement receives both base and spectrum:
\begin{equation}
\begin{aligned}
h_{\rm bc}&=\operatorname{GELU}
 \bigl(W_B\mathrm{LN}_{\rm aff}(B^\circ)
       +W_s\mathrm{LN}_{\rm aff}(s)\bigr),\\
z_{\rm bc}&=\mathrm{LN}_\epsilon(B^\circ+W_u h_{\rm bc}).
\end{aligned}
\label{eq:supp-base-conditioned-direct}
\end{equation}
Its rank-32 mapper/total counts are 275,024/299,606. Mapper capacity is within 1.4\% of base-fine; total counts differ because of auxiliary heads. It uses Eq.~\eqref{eq:supp-direct} with its own state and alignment weight, without base projection, norm control, fine loss or explicit zero-scale bypass.

\begin{table*}[!tp]
\centering\small
\caption{Complete fixed-epoch curves. (a) Retargeting varies alignment $\alpha$; base-fine varies injection $q$, with cached reference-fixed logit probes. (b) Validation-high-anchor base-conditioned replacement versus selected base-fine. (c) Common-$\alpha$ curves, distinct from per-seed selection. Higher recognition/cosine and lower KL are favorable.}
\label{tab:supp-frontier}\label{tab:supp-base-conditioned-selected}\label{tab:supp-base-conditioned-frontier}
\textbf{(a) Spectrum-only alignment and injection}\par\smallskip
\input{tables/tab_full_frontier}
\par\medskip\textbf{(b) Base-conditioned selected comparison}\par\smallskip
\input{tables/tab_base_conditioned_direct}
\par\medskip\textbf{(c) Base-conditioned alignment grid}\par\smallskip
\input{tables/tab_base_conditioned_frontier}
\end{table*}
\begin{figure*}[!tp]
\centering
\includegraphics[width=\textwidth]{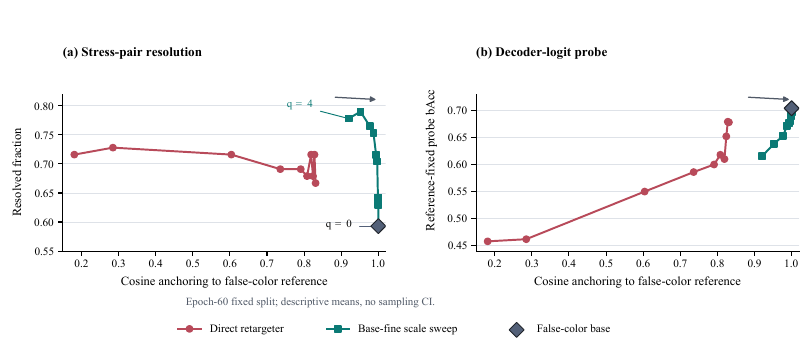}
\caption{Epoch-60 fixed-split anchoring--retention sensitivity. (a) Spectrum-only retargeting and base-fine on 27 stress pairs. (b) The shared reference-fixed decoder-logit probe on 461 objects, using cached logits with reference bAcc $0.704$. Curves are descriptive three-seed means over observed ranges; pair response is nonmonotone. Selected base-fine gives pair resolution $0.778$ at reference cosine $0.939$.}
\label{fig:supp-fixed-frontier}
\end{figure*}
Base-conditioned direct/base-fine gives cosine/category/pair/proxy $0.977/0.813/0.605/0.852$ versus $0.939/0.905/0.778/0.938$. Direct-minus-base-fine session intervals are $[-0.165,-0.043]$ for category, $[-0.184,-0.031]$ for proxy and $[-0.377,-0.006]$ for pair. Reference-fixed decoder bAcc is $0.590/0.620$ and KL $2.029/1.400$; their bAcc interval spans zero. These results characterize the tested architecture, alignment grid and validation-high-anchor selection.

\subsection{Identical-Residual and Matched Ablations}
Own-token addition and pooled replacement use the same spectra, residual, target support, selected CE checkpoints $(3,6,4)$ and $q=4$. Both arms use the common recomputed decoder configuration. Category/proxy bAcc is $0.896720/0.892346$ and $0.930654/0.924114$; reference-fixed decoder bAcc is $0.700684/0.589521$, and refitted bAcc $0.790621/0.741123$.
\begin{table*}[!tp]
\centering\small
\caption{Receiving-token composition matters at closely matched recognition. (a) Three optimization seeds with identical residuals/checkpoints/scales; KL P95 uses the jointly recomputed reference. (b) Paired acquisition-session intervals on 461 objects in 66 sessions; all contrasts are exploratory. Positive bAcc and negative KL differences favor the first condition.}
\label{tab:supp-composition-decoder-seeds}\label{tab:supp-composition-decoder-intervals}
\textbf{(a) Identical-residual results by seed}\par\smallskip
\input{tables/tab_decoder_same_residual_seeds}
\par\medskip\textbf{(b) Paired composition and input-code contrasts}\par\smallskip
\input{tables/tab_decoder_contrast_intervals}
\end{table*}
Intervals use 10,000 shared resamples of all objects within each test session (seed 20261004), recomputing each seed's metric and averaging paired differences. A second reference-trained head, fitted on cached logits and applied unchanged to both recomputed arms, retains a fixed-readout gain of $0.110576$ [$0.045640,0.163843$].

The four-cell injection comparison returns exact $G$ for no-injection addition and $R(B^\circ)$ for no-injection replacement. The fixed-readout addition advantage is $0.097731$ without injection and $0.111163$ with injection; the interaction is $0.013432$ [$-0.014300,0.058071$]. Interactions under the second fixed head and refitted heads also span zero. Category injection gains remain positive in both operators: $0.096165$ [$0.055093,0.158895$] and $0.104924$ [$0.048272,0.209174$]. Thus general grid preservation and content-specific transfer are distinct effects.
\begin{table*}[!tp]
\centering\small
\caption{Composition and supervision controls. (a) Same-residual four-cell comparison with and without injection in the jointly recomputed runtime. Fixed preserves the reference-trained decoder readout. (b) Complete final-epoch matched ablation with refitted cached-logit probes; Pair is both-endpoint correctness, Category/Proxy state are pooled bAcc and Cosine is reference anchoring. Zero scale is a reference endpoint.}
\label{tab:supp-composition-injection}\label{tab:supp-matched-composition}
\textbf{(a) Composition crossed with injection}\par\smallskip
\input{tables/tab_composition_injection_supp}
\par\medskip\textbf{(b) Matched representation and supervision ablation}\par\smallskip
\input{tables/tab_compact_composition}
\end{table*}

Raw-plus-NMF readout is a non-VLA-facing 136-D feature. Its frozen-decoder counterpart is a label-free $136\to4096$ reduced-QR isometry (seed 20260710) followed by row LayerNorm and replacement. Random-label adapters permute training/validation labels with seeds $s+991/s+1991$, preserving class counts, losses and endpoint weights. They retain epoch 60 and inherit correct-fine scale; external readouts/test use true labels. No-pair weighting sets all $\omega_i=1$. No-fine loss removes only its auxiliary term and selects its own scale.

The raw/NMF input ablation pads inactive coordinates to retain 136 inputs and 320,348 parameters. It evaluates the pooled surrogate $T_q$, not a full injected grid, uses 74 training pairs at threshold $0.80$, minimum validation-CE selection and the central 27 test pairs. Raw-plus-NMF selects epochs $(4,1,4)$ and $q=(4,4,4)$. Only these three rows constitute its matched input comparison.
\begin{table*}[!tp]
\centering\small
\caption{Input and execution calibration. (a) Capacity-matched raw/NMF inputs, with active width reported separately; Pair$_{27}$ uses the central stress pairs. (b) Validation-CE checkpoint states. (c) Validation mean-displacement allowances; reported tails are measured on test. Decoder values in (b,c) use cached-logit sensitivity, separate from joint recomputation.}
\label{tab:supp-raw-nmf}\label{tab:supp-validation-frontier}\label{tab:supp-common-budget}
\textbf{(a) Spectral input ablation}\par\smallskip
\input{tables/tab_raw_nmf_ablation}
\par\medskip\textbf{(b) Validation-selected checkpoint states}\par\smallskip
\input{tables/tab_validation_checkpoint_frontier}
\par\medskip\textbf{(c) Common displacement allowances}\par\smallskip
\input{tables/tab_r6_common_budget}
\end{table*}

\subsection{Achieved Movement and Readout Sensitivity}
At allowance $0.2$, base-fine/no-projection selects $(1.5,1.5,1)/(1.5,1.5,1.5)$ with validation mean displacement $(0.1615,0.1615,0.1084)/(0.1621,0.1621,0.1619)$. Selection uses composed-head validation CE before utility and smaller-scale ties. The unprojected arm changes only base projection.

Bisection can instead match achieved validation movement. At target $0.15$ within $10^{-5}$, base-fine/no-projection test means are $0.147383/0.147391$ and category bAcc $0.870742/0.871170$. Mean/P95 KL is $0.057333/0.163369$ versus $0.055741/0.160344$; paired differences are $0.001592$ [$0.000840,0.002378$] and $0.003025$ [$-0.003099,0.015307$] (10,000 shared 66-session resamples, seed 20261008). This matched-movement result separates projection's attribution role from calibration of the achieved state.

Seven fixed readout transforms test pooling dependence after representation selection: native width, five signed matrices (seeds 20260712--20260716) and contiguous-block means. Across 21 seed--transform combinations, category/proxy gains average $0.102/0.103$, with ranges $0.071$--$0.132$ and $0.071$--$0.127$.
\begin{table}[!tbp]
\centering\small
\caption{Readout robustness after fixing the representations. Mean [minimum, maximum] bAcc spans seven predeclared transformations and three training seeds; ranges are not sampling confidence intervals.}
\label{tab:supp-readout-definition}
\input{tables/tab_readout_definition_robustness}
\end{table}

\subsection{Residual Constraints and Budget Sensitivity}
The component ablation changes only projection or norm control, with rank 64, three seeds, 60 epochs and validation-only scale/readout selection. Norm budgets are diagnostic settings, not test-selected choices. Removing the bound raises the pre-scale/effective residual ratio to $0.671/1.333$, lowers cosine to $0.655$ and raises KL to $7.148$. Removing orthogonalization retains norm control and can improve recognition. Projection defines a base-complement attribution; norm and scale control its magnitude.
\begin{table*}[!tp]
\centering\small
\caption{Constraint sensitivity. (a) Fixed-split final-epoch budget/component ablation with refitted cached-logit probes. Ratios distinguish pre-scale $\|D\|/\|B\|$ and injected $q\|D\|/\|B\|$; displayed $q$ averages validation-selected scales. (b) Separate 5/15-epoch nested constraint study; Sel. counts inner-validation choices, and the decoder probe is reference-fixed.}
\label{tab:supp-constraints}\label{tab:supp-nested-constraints}
\textbf{(a) Components and norm budget}\par\smallskip
\input{tables/tab_constraint_ablation}
\par\medskip\textbf{(b) Session-held-out constraint selection}\par\smallskip
\input{tables/tab_nested_constraint_confirmation}
\end{table*}
\begin{figure*}[!tp]
\centering
\includegraphics[width=0.92\textwidth]{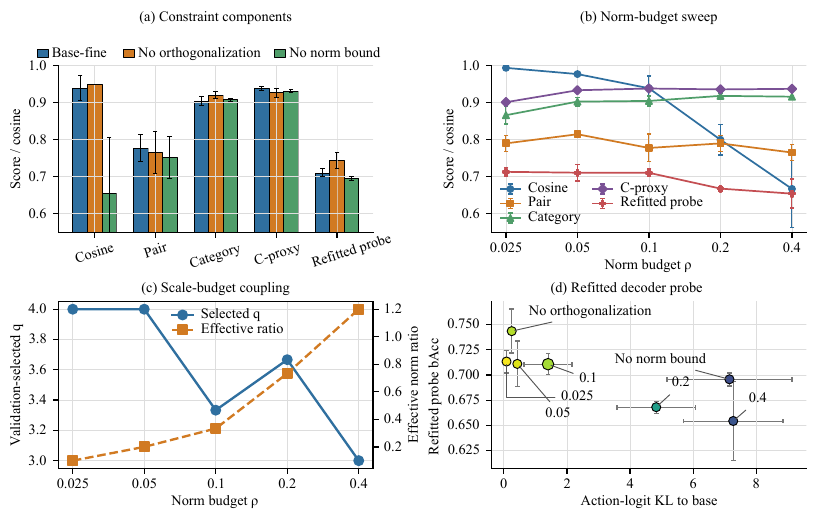}
\caption{Image-disjoint constraint sensitivity at the final epoch, using cached logits and refitted decoder probes (Table~\ref{tab:supp-constraints}a). Bars compare $\rho=0.1$, no orthogonalization and no norm bound. The budget sweep separates recognition, anchoring and accessible decoder information; selected scale partly compensates for budget. Error bars are SD across three seeds.}
\label{fig:supp-constraints}
\end{figure*}

The separate nested constraint study fixes rank 64/fine weight 1, uses five selection and 15 refit epochs, and compares orthogonal budgets $\{0.025,0.05,0.1,0.2,0.4\}$ with no projection at $0.1$. Inner validation requires cosine $\geq0.95$ before utility selection; every fold refits preprocessing on outer training. It chooses orthogonal variants in 77/86 folds. The selected state has cosine/category/proxy $0.980/0.889/0.924$ and reference-fixed probe bAcc $0.705$ versus $0.673$ for fixed orthogonal $\rho=0.1$: difference $0.032$ [$0.009,0.061$]. Category/proxy contrasts and all contrasts with unprojected $\rho=0.1$ span zero. This study is distinct from principal 12/60 recognition and the separate short-budget decoder diagnostic.
\begin{figure*}[!tp]
\centering
\includegraphics[width=\textwidth]{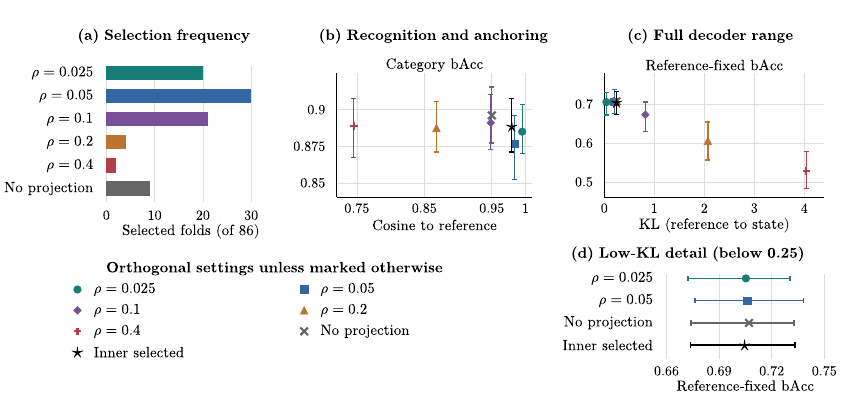}
\caption{Nested constraint diagnostic: five selection and 15 refit epochs. (a) Inner-validation choices. (b,c) Recognition--anchoring and reference-fixed probe--movement, with session-clustered 95\% intervals. (d) Overlapping low-KL settings separated on rows; estimates and intervals are unchanged. Selected constraints retain high anchoring without large-budget probe degradation.}
\label{fig:supp-nested-constraints}
\end{figure*}

\section{Session-Held-Out Generalization}
\label{sec:supp-nested60}
Each of 86 outer folds holds out one complete session. From the remaining 85, 48 candidate inner splits use validation fraction $0.16$ and seeds $20260809+f+o$. Splits retain all six categories and minimize category-proportion $L_1$ drift plus $0.1$ times size error. Inner training fits spectral transforms and candidates; outer training refits both after selection. Deterministic frozen visual features contain no population-fitted statistic, so they can be reused across folds.

Inner ambiguity thresholds are $\{0.80,0.85,0.90,0.95\}$. The strictest cutoff with at least 20 validation pairs is preferred; otherwise maximum count and then the strictest cutoff. All folds use the $0.80$ fallback, with 1--8 validation pairs (mean 5.27). Their fold-specific training pairs receive endpoint weight 4. Candidate/refit seeds are $20260809+f$ and $20260809+10000+f$, with 12/60 epochs.

Base-fine and no-projection vary rank $\{32,64\}$, fine-loss weight $\{0.5,1,2\}$ and the ten injection scales. Ridge 1 supplies validation utility $\mathrm{bAcc}+0.25\,\mathrm{Pair}$; ties prefer cosine, then smaller scale, rank and fine weight. This gives 60 evaluated operating points per residual family, sharing fits across scales. Base-conditioned direct varies rank $\{32,64\}$ and $\alpha\in\{0,0.01,0.05,0.1,0.5,1,10\}$ (14 candidates). It requires cosine $\geq0.94$, then orders by utility, cosine, smaller $\alpha$ and rank; if none qualifies, cosine precedes utility. Final category/proxy ridge regularization is selected on inner validation after fixing the representation.

Base-fine selects rank 32/64 in 39/47 folds; scales $0.75/1/1.5/2/3/4$ in 2/3/8/6/18/49; and fine weight $0.5/1/2$ in 23/29/34. Direct selects ranks in 38/48 folds and $\alpha=0/0.01/0.05/0.1/0.5/1$ in 1/2/50/20/5/8. All 85 outer-training sessions refit preprocessing, adapter and auxiliary heads; every held-out object contributes one prediction.

\begin{table*}[!tp]
\centering\small
\caption{Recognition gains across unseen acquisitions. (a) Principal 12/60-epoch comparison on the same 2,291 out-of-fold objects; category macro-F1 and equal-weight session summaries expand the main results. Session means/medians average present-class recalls; Worst-10 averages each method's ten lowest sessions. (b) Category support and recall. Gray identifies SPECTRA-VLA; bold marks favorable means, not significance.}
\label{tab:supp-session-summary-expanded}\label{tab:supp-oof-classes}
\textbf{(a) Aggregate and session-level recognition}\par\smallskip
\input{tables/tab_session_summary_expanded}
\par\medskip\textbf{(b) Category support and recall}\par\smallskip
\input{tables/tab_r6_oof_classes}
\end{table*}

Base-fine raises pooled category bAcc from $0.812$ to $0.883$, with paired session gain interval $[0.050,0.107]$. Deleting each session from the fixed predictions leaves gains $0.0680$--$0.0779$ over the base and $0.0651$--$0.0769$ over direct. These ranges describe influence conditional on the trained folds.

Principal intervals use 10,000 session-bootstrap draws (seed 20260891), recomputing pooled macro recall over the fixed six-category/four-state class set. An absent class contributes zero recall; all classes occur in every principal draw. Resampling conditions on realized fits and does not repeat training/selection. Optimization SD and session uncertainty therefore have different interpretations.

\paragraph*{Selection-utility sensitivity}
Removing the pair term changes 11/86 base-fine choices, 7 no-projection and 25 direct choices. The 11 changed base-fine folds repeat inner fitting/readout selection and 60-epoch outer refitting; the other 75 predictions remain fixed. Category bAcc becomes $0.881547$ versus $0.882859$, difference $-0.001313$ [$-0.006327,0.001768$]; proxy bAcc becomes $0.917447$ versus $0.916076$, difference $0.001371$ [$-0.000116,0.003633$]. Training-pair weights and every other setting remain unchanged. Changed choices of the other methods are not refitted.

\section{Ambiguity and Clustered Inference}
\label{sec:supp-pairs}
Candidate pairs have different categories and images, with cosine above a validation-selected threshold in target-pooled \emph{paired RGB} tokens. No spectral-distance filter is used. Sorting is by decreasing cosine; ties retain deterministic enumeration order. Greedy matching makes object endpoints disjoint. Pair $p=(i,j)$ is resolved for seed $s$ only when both endpoints are correct:
\[
R_{sp}=\mathbb 1[\hat y_{si}=y_i\ \wedge\ \hat y_{sj}=y_j].
\]
Resolution averages this indicator over the inventory. Hard bAcc is macro recall over its endpoints. Method differences are averaged over three seeds before resampling, so seeds add no sampling units.

The fixed-split grid $\{0.75,0.80,0.85,0.90,0.95\}$ uses the strictest cutoff yielding at least 20 validation pairs, else maximum count and then the strictest cutoff. It selects $0.75$ with 24 validation pairs, producing 27 test pairs: 54 distinct objects, 49 images and 38 sessions. Endpoint records contain category, image/component identity, RGB cosine and spectral angle. Repeated images/sessions make this an object-disjoint \emph{clustered stress sample}.

Primary pair inference uses 100,000 session-node product-multiplier draws (seed 20260713). Each session receives one independent $\mathrm{Exp}(1)$ weight. A cross-session pair receives the product of both endpoint weights; a within-session pair receives that weight once. The weighted seed-averaged paired difference is recomputed each draw. Base-fine-minus-base gives point estimate $+0.185$ and session interval $[-0.030,0.443]$. Image-node, delete-session jackknife and pair-resampling sensitivities are $[-0.028,0.468]$, $[-0.068,0.438]$ and $[0.037,0.346]$. The point estimate is reported alongside clustered uncertainty; category/session evidence provides the supported recognition gain.

\begin{table*}[!tp]
\centering\small
\caption{Ambiguity-definition sensitivity. (a) Threshold-specific inventories; resolution averages seed estimates, and exact-CI envelopes span per-seed intervals rather than providing pooled coverage. (b) Continuous similarity over 215 fixed object-disjoint pairs, with differences averaged across seeds before inference. Its session-node intervals use 20,000 draws (seed 20260805).}
\label{tab:supp-thresholds}\label{tab:supp-continuous-similarity}
\textbf{(a) Threshold sensitivity}\par\smallskip
\input{tables/tab_ambiguity_thresholds}
\par\medskip\textbf{(b) Continuous RGB-token similarity}\par\smallskip
\input{tables/tab_similarity_continuous}
\end{table*}
\begin{figure*}[!tp]
\centering
\includegraphics[width=0.96\textwidth]{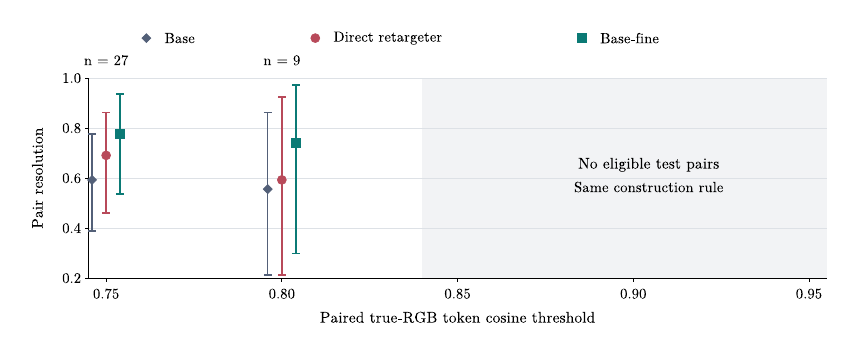}
\caption{Threshold sensitivity with separately reconstructed inventories: $0.75$ yields 27 pairs, $0.80$ yields 9, and stricter thresholds yield none. Points are three-seed means; whiskers span per-seed exact pair-binomial intervals and are descriptive rather than a pooled 95\% interval.}
\label{fig:supp-threshold-sensitivity}
\end{figure*}
The continuous inventory has 215 disjoint cross-category pairs above cosine zero, 155 images and 65 sessions, grouped into four predefined quantiles. Spearman association between RGB similarity and spectral gain is $0.115$ [$-0.073,0.308$]. Quartile means are positive but nonmonotone; this sensitivity does not select the central stress inventory.

\begin{table}[!tbp]
\centering\small
\caption{Paired differences (S6). Fixed-split decoder/prototype rows: epoch-60 representation-refitted readouts, pooled scores. Nested rows: reference-fixed readouts, equal-weight session means (not pooled OOF). Accuracy up/KL down favors base-fine; bold excludes zero. Base/direct contrasts are primary/exploratory. Direct-pair CIs are descriptive; other CIs use indicated image/session units. Seed SD: optimization spread.}
\label{tab:supp-paired-differences}
\input{tables/tab_paired_differences}
\end{table}

\section{Decoder-Readout Diagnostics}
\label{sec:supp-short-action}
\subsection{Fixed-Split Recognition and Decision-Function Reuse}
The generic sorting prompt contains no category/state word and the supplied mask specifies the target. All representations use the same 256 action coordinates and ridge family. Refitted probes measure accessible four-state information; reference-fixed probes preserve the established standardizer/head. The category-lookup arm predicts six categories from the \emph{same frozen logits}, then applies Table~\ref{tab:supp-material-state-map}.

\begin{table*}[!tp]
\centering\small
\caption{Fixed-epoch decoder readouts using cached logits. (a) Refitted and category-lookup readouts, plus one reference-trained ridge with regularization 10 applied unchanged. The combined column reports its four-state bAcc, then native action-symbol argmax agreement with the false-color reference over 256 logits. (b) Proxy recall after category prediction and lookup. Means use three seeds; higher bAcc/agreement and lower reference-to-altered KL are favorable.}
\label{tab:supp-action-aggregate}\label{tab:supp-shared-base-probe}\label{tab:supp-material-lookup-recall}
\textbf{(a) Complete decoder comparison}\par\smallskip
\input{tables/tab_primary_action}
\par\medskip\textbf{(b) Category-to-proxy recall}\par\smallskip
\input{tables/tab_material_lookup_state_recall}
\end{table*}

Epoch-60 cached logits give refitted base-fine/base bAcc $0.711/0.704$ (session difference $[-0.043,0.095]$); reference-fixed bAcc is $0.620$, difference $-0.084$ [$-0.137,-0.030$]. Jointly recomputing the same residuals at $q=(2,4,4)$ gives $0.617358/0.694685$, difference $-0.077327$ [$-0.117507,-0.035220$]. Earlier CE-selected epochs $(3,6,4)$ yield $0.701$ at $q=4$: checkpoint and scale differ, but the recomputed runtime and reference-fixed readout are identical. Cached sensitivities remain separate.

\subsection{Session-Shift Diagnostic}
The separate five-selection/15-refit-epoch protocol uses the same 2,291 objects and 86 outer sessions. Inner split seeds are $20260710+k+o$ for 48 trials; splits retain all categories and minimize proportion drift plus $0.1$ size error. Candidates vary ranks $\{16,32,64,128\}$, fine weights $\{0.5,1,2\}$, the full scale grid and direct alignment grid. Ridge 1 supplies category-plus-$0.25$-pair utility; exact ties prefer smaller rank, scale, representation-readout value, fine weight and alignment weight. Final regularization follows representation selection. All transforms are inner-trained, then outer-refitted; the action head is fitted only on base logits and fixed across representations.

\begin{table*}[!tp]
\centering\small
\caption{Short-budget session-held-out reference-fixed decoder diagnostic. Each representation is inner-selected, refitted for 15 epochs and tested once per held-out object. Pooled OOF bAcc uses 2,291 predictions; equal-weight session summaries average present-class recalls. This five/15-epoch study differs from principal 12/60 recognition and the fixed-rank constraint experiment.}
\label{tab:supp-nested-fixed-coordinate}
\input{tables/tab_nested_loso}
\end{table*}
Base-fine/base OOF bAcc is $0.684/0.705$. The mean paired session difference is $-0.024438$, with 4,000 session-bootstrap interval $[-0.049086,0.001150]$ (seed 20260710), rather than the pooled-score difference $-0.020785$. Table~\ref{tab:supp-paired-differences} rounds these bounds outward to three decimals. Session sizes range from 1 to 122 objects (median 19); instance weighting gives approximately $-0.022$ versus $-0.024$ unweighted. This is a fixed-budget shift diagnostic.

\subsection{Normalization and Action-Coordinate Sensitivity}
A raw-token-trained ridge selects regularization 100 on raw validation logits, then stays fixed on raw, normalized $G$ and injected states. The $G$-trained reference selects 10. State-specific refits use the common grid.
\begin{table*}[!tp]
\centering\small
\caption{Decoder-coordinate decomposition and paired uncertainty. (a) Fixed versus state-specific readouts on identical objects/coordinates. (b) Conditional paired intervals for OpenVLA coordinate changes and SpatialVLA gains; 10,000 shared 66-session draws (seed 20260922), with absent-class recall zero. Bold marks intervals excluding zero, including adverse effects.}
\label{tab:supp-r7-coordinate}\label{tab:supp-r7-intervals}
\textbf{(a) Normalization and injection}\par\smallskip
\input{tables/tab_r7_coordinate_decomposition}
\par\medskip\textbf{(b) Paired session differences}\par\smallskip
\input{tables/tab_r7_paired_intervals}
\end{table*}
The raw-trained readout changes from $0.730378$ to $0.647454$ after normalization and $0.610655$ after injection. Individual differences span zero; the total raw-to-injected interval is $[-0.205931,-0.006330]$. The $G$-trained fixed readout declines by $0.084164$. Seventeen bootstrap draws omit a class, handled identically in both arms.

The separate normalized-input vocabulary diagnostic clips each original band by imagewise 2nd/98th percentiles before interpolation. Bands $[24,60,96]$ are normalized again for false color; the object spectrum is L2-normalized and train-standardized without NMF. Float16 mask weights use area occupancy divided by its maximum plus $10^{-8}$. The 319,292-parameter model uses threshold $0.80$, CE-selected epochs $(1,2,3)$ and $q=4$. All 32,064 logits arise from one forward before extracting action coordinates and three preset 256-coordinate complements.
\begin{table*}[!tp]
\centering\small
\caption{Separate normalized-input vocabulary diagnostic (Section~S6.3). Action-coordinate movement versus three preset vocabulary complements at $q=4$. Each 256-coordinate subset is internally normalized from the same full-vocabulary forward. Objectwise means across adapter seeds precede paired acquisition-session inference. Complement sets are descriptive controls rather than a vocabulary-wide null.}
\label{tab:supp-random-vocab}
\input{tables/tab_random_vocab_specificity}
\end{table*}

\section{Numerical Geometry and Forward Equivalence}
\subsection{Executed-Token Geometry}
The normalized false-color reference $G$ is the exact $q=0$ endpoint: its token/logit differences and KL are zero, top-1 agreement is one and downstream differences are zero. Positive-scale projection is numerically, rather than exactly, orthogonal. Across 6,885 object states, maximum absolute normalized dot product is $2.49\times10^{-7}$ and maximum norm-ratio error from $\rho=0.1$ is $1.13\times10^{-6}$ (means $4.01\times10^{-8}$ and $3.77\times10^{-7}$). Distinct reference pooling routines differ by at most $9.54\times10^{-6}$ (mean $6.70\times10^{-8}$); this is separate from the bypass identity.

For real-arithmetic mean-centered $G_t$, define $D_c=D-\bar D\mathbf1$,
\[
\alpha_t=\frac{G_t^\top D_c}{\|G_t\|^2},\qquad
\beta_t=\frac{\|D_c\|^2}{\|G_t\|^2}.
\]
For nonzero inputs, LayerNorm's positive scale cancels from cosine:
\[
\cos(G_t,\mathrm{LN}_\epsilon(G_t+qD))
 =\frac{1+q\alpha_t}{\sqrt{1+2q\alpha_t+q^2\beta_t}}.
\]
Pooled orthogonality does not set every $\alpha_t$ to zero. With
$c_t=[d^{-1}\|G_t+qD_c\|^2+\epsilon]^{-1/2}$,
$a_t=1-m_t+m_tc_t$ and $b_t=m_tc_tq$, fractional-mask composition gives
\[
\cos(G_t,a_tG_t+b_tD_c)
 =\frac{a_t+b_t\alpha_t}
        {\sqrt{a_t^2+2a_tb_t\alpha_t+b_t^2\beta_t}}.
\]
Token alignment and occupancy jointly determine executed movement; decoder KL is measured empirically.

\subsection{Training Surrogate and Executed Pool}
At epoch 60, $q=(2,4,4)$ for seeds $(0,1,2)$, the 461 test objects yield 1,383 training/executed-pool comparisons: cosine 0.9977, mean/P95 relative distance $\|T_q-C_q\|/\|T_q\|=0.0569/0.1328$, and coordinate-mean residual energy 0.0098\%. Validation occupancy cuts $0.0204,0.0522,0.1080$ give quartile distances $0.0537,0.0276,0.0497,0.0965$ and cosines $0.9981,0.9994,0.9985,0.9948$.
The unchanged auxiliary category head is evaluated separately at validation-CE epochs $(3,6,4)$ and $q=4$: prediction disagreement is $2.82\%$, with mean true-category margin changes $-0.042,0.205,0.174$ by seed. Thus small aggregate geometric differences need not imply identical decisions.
\begin{table}[!tbp]
\centering\small
\caption{Auxiliary-head behavior by validation occupancy quartile (validation-CE epochs $3,6,4$; $q=4$). Margins are true-category logits minus the largest competitor, reported as $T_q/C_q$. Section S7.2 separately gives epoch-60 geometry.}
\label{tab:supp-map-diagnostic}\label{tab:supp-r7-head}
\input{tables/tab_r7_head_occupancy}
\end{table}
Protocol normalization changes the raw decoder response: raw-to-$G$ action KL mean/P95/P99 is $2.610/5.413/6.888$, reverse KL $2.399$ and top-1 agreement $0.163$. Operational logits agree with the cached reference on all 2,295 objects. The explicit zero endpoint is $G$.

\subsection{Official Forward and Token Reinsertion}
On 171 unmodified held-out RGB observations, the official multimodal OpenVLA forward and extraction/reinsertion of its \emph{same raw visual grid} use identical prompt, masks, positions, pixels, weights and bfloat16 inference. Across all 32,064 runtime coordinates, maximum absolute logit difference, full-vocabulary KL and action-mass difference are zero; full/action-subset top-1 agreements are one. Mean action-subset mass is $0.999966$. This validates the visual insertion interface before normalization and readout; the method's zero endpoint is the protocol-normalized false-color reference.

\section{Backbone and Observation Robustness}
\subsection{Second Frozen VLA}
SpatialVLA repeats the same objects, train-only features, rank 64, three seeds, 60 epochs and validation scale/ridge selection. Pooled heads refit per representation; correspondence controls retain correct-spectrum heads. One action ridge fitted/selected on base training/validation logits stays fixed throughout.
\begin{table*}[!tp]
\centering\small
\caption{SpatialVLA replication. (a) Matched states with refitted pooled heads and one reference-fixed decoder probe. KL/top-1 use all 8,194 action coordinates. (b) Correspondence controls retain correct-spectrum pooled heads; the action head remains reference-fixed. At zero scale the grid/logits equal the base, though pooled scores use different heads.}
\label{tab:supp-spatialvla}\label{tab:supp-spatialvla-controls}
\textbf{(a) Matched representations}\par\smallskip
\input{tables/tab_spatialvla}
\par\medskip\textbf{(b) Spectral correspondence}\par\smallskip
\input{tables/tab_r2_spatialvla_controls}
\end{table*}
Base-fine reaches pooled cosine $0.985$, with category/proxy bAcc $0.914/0.947$ versus base $0.812/0.871$. Reference-fixed decoder performance remains lower, reproducing the distinction between enriched token information and retained decoder decisions.

\subsection{Grounding}
Every supplied region re-encodes the masked false-color image and supplies its mean spectrum. Empty regions give black images, zero spectra and masks. Each representation's readout is trained on ground-truth-region logits and fixed across test perturbations, as is its scale.

The predicted-mask model is pretrained FCN-ResNet50 trained for six epochs on $160^2$ RGB with ImageNet normalization, random flips, AdamW $8\times10^{-5}$/$10^{-4}$, inverse-square-root pixel-frequency weights (background $\times0.25$) and auxiliary CE weight $0.35$. Foreground mIoU on 80 fixed validation images (seed 4137) selects the checkpoint. Resized test logits are argmaxed. Given the target category, components below 40 pixels are removed and the largest remaining component is chosen, without matching the target. The separate oracle-matched diagnostic uses maximum target IoU.
\begin{table}[!tbp]
\centering\small
\caption{Grounding sensitivity with fixed ground-truth-region-trained decoder readouts and scales. Every condition re-encodes its region. IoU compares with the target mask; predicted masks receive the target category and select its largest predicted component.}
\label{tab:supp-grounding}
\input{tables/tab_grounding}
\end{table}
\begin{figure*}[!tp]
\centering
\includegraphics[width=0.97\textwidth]{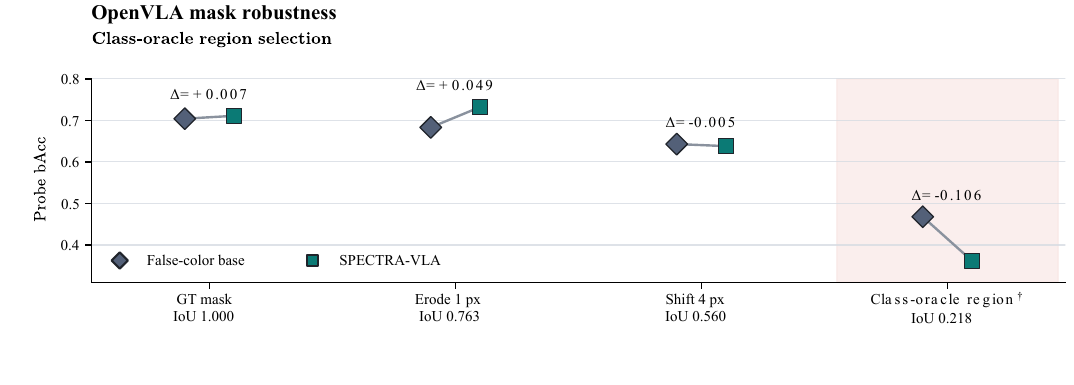}
\caption{Region re-encoding on 461 objects in the earlier cached decoder configuration (Section S1.6). Each representation's ground-truth-mask-trained probe and scale remain fixed across masks. Probes use 256 first-position coordinates; points are descriptive means without sampling intervals. Erosion preserves base-fine/base bAcc $0.732/0.683$; largest predicted components supplied with the target category ($^\dagger$) give IoU $0.218$ and bAcc $0.362/0.468$.}
\label{fig:supp-mask-robustness}
\end{figure*}
A four-pixel shift gives approximately neutral $0.638/0.643$. Moderate region errors and poor target selection therefore have different effects.

\subsection{Capacity and Spectral Bands}
The raw-only rank sweep uses 120 bands without NMF. Parameter count is $2p+r(p+1)+4096r+2\cdot6(4096+1)$, or $4217r+49404$ for $p=120$. Raw-only rank 64 has 270,128 mapper/319,292 total parameters, versus the principal 136-input model's 271,184/320,348. Category/pair at rank 64 is $0.894/0.796$, versus $0.884/0.778$ at rank 256.

Session-held-out band exclusion fixes rank 64, fine weight 1 and $\rho=0.1$, with 12/60 epochs. Inputs are full $[0,\ldots,119]$, base $[119,60,0]$, base exclusion $[1,\ldots,59,61,\ldots,118]$, random $[20,75,82]$ and contiguous $[30,31,32]$. Selected values fill a zero-padded 120-channel vector. NMF uses $\min(16,\text{selected bands})$ components, padded to 16 after training-fitted scaling/standardization. The full-input category bAcc is $0.886$ versus base-band $0.835$, difference $0.050$ [$0.032,0.075$]; exclusion retains $0.878$. The fixed hyperparameters distinguish this control from principal nested selection.
\begin{table*}[!tp]
\centering\small
\caption{Capacity and spectral-input sensitivity. (a) Raw-only rank sweep, three seeds, validation-selected scales; counts exclude frozen VLA/readouts. (b) Session-held-out band exclusion with unchanged false-color tokens; only the fine-path input changes. Higher category/proxy/pair/cosine is favorable.}
\label{tab:supp-rank}\label{tab:supp-band-exclusion}
\textbf{(a) Rank and parameter count}\par\smallskip
\input{tables/tab_rank_capacity}
\par\medskip\textbf{(b) Band exclusion}\par\smallskip
\input{tables/tab_r2_band_exclusion}
\end{table*}

\subsection{Rendering and Radiometry}
\label{sec:supp-six-render}
Six-render augmentation compares HSI display constructions: fixed sensor $[119,60,0]$, percentile-scaled low/mid/high SWIR $[119,60,0]$, NIR $[16,8,0]$, random $[105,61,7]$, gain/offset $[96,60,24]$ and train-fitted PCA. Per training image, 512 sampled pixels (seed 20260710) determine 2nd/98th percentiles $L_b,H_b$. Display is $255\operatorname{clip}[(x_b-L_b)/(H_b-L_b+10^{-6}),0,1]$, except fixed sensor uses $(0,65535)$. Gain/offset maps $x_b$ to $(1+0.20\sin\phi_b)x_b+0.05\max(H_b-L_b,10^{-6})\cos\phi_b$, $\phi_b=2\pi b/119$. PCA fits three randomized-SVD components on the same sampled pixels and their training display percentiles.

Each epoch shuffles 8,352 rows (six per training object) with repeated spectra/labels and $\omega_i=1$. Mean validation category bAcc plus $0.25$ mean pair resolution selects $q=(4,4,3)$, with smaller-scale ties. One readout trained on all augmented rows and each seed's scale remain fixed across test views. Paired-RGB tokens define the visually similar stress pairs; the six training and test views above are HSI renderings.
\begin{table}[!tbp]
\centering\small
\caption{Six-render augmentation under one fixed scale/readout per seed, with three-seed category/pair means. All test views use the same 461 objects and 27 pairs; this is separate from unaugmented primary evaluation.}
\label{tab:supp-shift}
\input{tables/tab_base_shift}
\end{table}
\begin{figure*}[!tp]
\centering
\includegraphics[width=\textwidth]{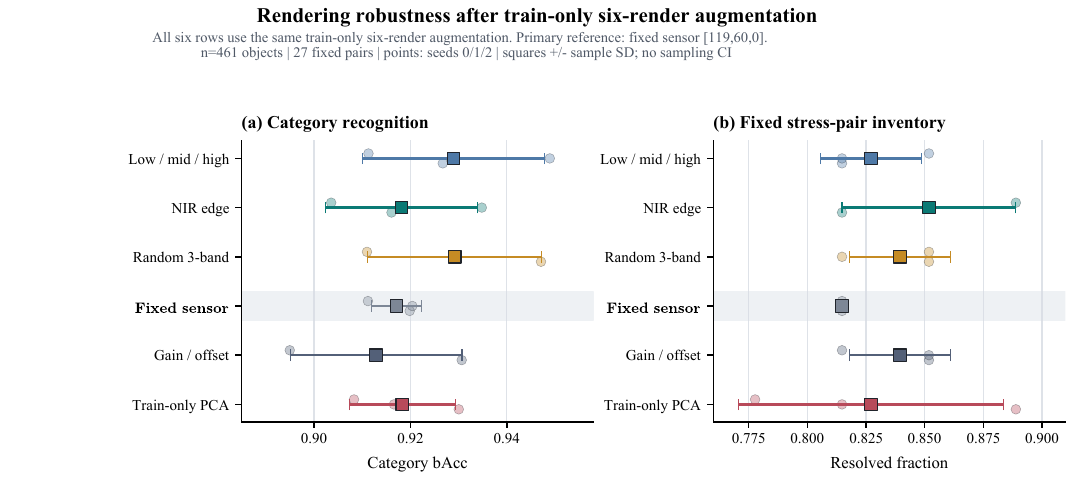}
\caption{Rendering and radiometric robustness after six-view training. Fixed sensor denotes the primary reference bands $[119,60,0]$. Category bAcc spans $0.913$--$0.929$ and pair resolution $0.815$--$0.852$. Points are seeds 0--2; squares/whiskers are means and seed SD, not sampling confidence intervals.}
\label{fig:supp-rendering}
\end{figure*}

\subsection{Class Errors and Conditional Recognition}
\label{sec:supp-failure}
Class recalls and confusion matrices use the same 461 objects and average per-seed row proportions rather than pooling seeded predictions. They separate six-category recognition, the four-state pooled proxy and the refitted decoder probe.
\begin{table}[!tbp]
\centering\small
\caption{Conditional recognition relative to the false-color base. Seed rows give aligned 461-object proportions; mean retention/correction includes acquisition-session intervals. $P(B+)$ denotes base correctness.}
\label{tab:supp-operational-linkage}
\input{tables/tab_operational_base_linkage}
\end{table}
\begin{table}[!tbp]
\centering\small
\caption{Class-resolved fixed-split recall, with methods compared in columns. Categories and proxy states follow Table~\ref{tab:supp-material-state-map}. Decoder-logit rows use refitted probes on frozen logits.}
\label{tab:supp-class-recall}
\input{tables/tab_class_resolved_recall}
\end{table}
\begin{table*}[!tp]
\centering\small
\caption{Row-normalized confusions (three-seed means), with methods side by side. Rows: true labels; columns: predictions. (a) Category IDs 1--6 follow Table~\ref{tab:supp-material-state-map}. (b,c) Its proxy states 0--3 are read from pooled states and refitted decoder-logit probes, respectively. Bold highlights diagonals, not cross-method best values; unrounded diagonals reproduce recall and bAcc.}
\label{tab:supp-material-confusion}\label{tab:supp-state-confusion}\label{tab:supp-fp-confusion}
\textbf{(a) Category recognition}\par\smallskip
\input{tables/tab_material_confusion}
\par\medskip\textbf{(b) Pooled proxy-state recognition}\par\smallskip
\input{tables/tab_state_confusion}
\par\medskip\textbf{(c) Decoder-logit probe}\par\smallskip
\input{tables/tab_fp_confusion}
\end{table*}

Conditional recognition partitions predictions by base correctness. If $p_s$ is base accuracy, $r_{sm}$ retention of base-correct predictions and $c_{sm}$ correction of base errors, then $\widehat{\mathrm{Acc}}_{sm}=p_s r_{sm}+(1-p_s)c_{sm}$. Base-fine retains $0.975$ of correct predictions and corrects $0.542$ of errors at $q=(2,4,4)$ and category ridge 100. The same identities/base predictions are shared with high-anchor direct; 100,000 66-session multiplier draws estimate paired uncertainty.
Main Fig.~\ref{main-fig:qualitative-pair} illustrates category correction, injection-induced harm, unresolved recognition and poor-region decoder failure. The first three show refitted category predictions, with fixed-readout outcomes reported separately. The fourth chooses the largest predicted target-category component and has IoU $0.0029$. These outcome-selected examples illustrate failure types rather than prevalence; display scaling and cropping do not alter the model inputs.

\section{Published Spectral Features and External Materials}
\subsection{Frozen Spectral Encoders}
Published spectral features test discrimination on the same objects \cite{spectralgpt,hypersigma,dofa,hypersl,maskedsst}. HyperSL uses the 120-band object mean with approximate 900--1700~nm coordinates (256-D output). MaskedSST uses 200 resampled bands in a masked $8\times8$ crop (96-D mean tokens); SpectralGPT uses 12 bands in a masked $96^2$ crop, zero mask ratio (768-D); DOFA uses masked $224^2$ crops at indices $[0,11,22,32,43,54,65,76,87,97,108,119]$ with micrometer wavelength coordinates (768-D); HyperSIGMA uses 120 bands in a masked $64^2$ crop (128-D). Raw spectra retain width 120.

HyperSL, MaskedSST and SpectralGPT have complete pretrained weights. DOFA's unavailable feature-normalization parameters retain identity initialization; its classification head is unused. HyperSIGMA lacks pretrained parameters in active attention blocks. These partial-load results form distinct diagnostic groups. All features use the same 1,392/442/461 objects, signed pooling, train-only standardization and validation-selected class-balanced ridge. Selected category coefficients are 1 (HyperSL), 0.1 (MaskedSST), 10 (DOFA), 1 (SpectralGPT), 10 (HyperSIGMA) and 0.01 (raw spectrum); proxy coefficients are selected independently. Pair resolution uses the same 27 stress pairs.
\begin{table*}[!tp]
\centering\small
\caption{Spectral separability and dataset-specific material recognition. (a) Published frozen features use the common training-only readout; Hard is hard-subset bAcc. Partial-load diagnostics ($^\dagger$DOFA, $^\ddagger$HyperSIGMA) are not ranked; $^\ddagger$also denotes out-of-domain pretraining. (b) Wood uses ten development splits and one class-balanced official test, so bAcc equals Acc.; SpecTex uses one within-source split without pairs. Fine width $k$ is validation-selected. $^\dagger$HyperSIGMA is partially loaded. Dim. is feature width and F1 is macro-F1.}
\label{tab:supp-foundation-results}\label{tab:supp-external}
\begin{minipage}[t]{0.49\textwidth}
\centering
\textbf{(a) Published spectral features}\par\smallskip
\input{tables/tab_foundation_peers}
\end{minipage}\hfill
\begin{minipage}[t]{0.49\textwidth}
\centering
\textbf{(b) External material recognition}\par\smallskip
\input{tables/tab_external_material}
\end{minipage}
\end{table*}
HSI-Adapter \cite{hsiadapter} supplies driving-segmentation code but had no released pretrained checkpoint when inspected on 22 September 2026, so its contribution is discussed in related work rather than numerically benchmarked.

\subsection{Dataset-Specific Material Transfer}
Mendeley Wood and SpecTex \cite{mendeleywood,spectex} test base-fine feature augmentation with a new basis fitted per dataset. These expanded ridge/PCA features are separate from native-width VLA composition.

Wood's 204-D three-band base joins means, standard deviations, 25th/75th percentiles and $8\times8$ block means. A standardized ridge with $\alpha=10$ predicts the 320-D mean spectrum; training-residual PCA supplies fine width $k\in\{0,4,8,16,32,64,128\}$. Validation Pair$+0.1$ bAcc selects $k$ and $\gamma\in\{0.25,0.5,1,2,4\}$ (zero for $k=0$), giving $k=4$ and 208 dimensions. The classifier uses standardized ridge 1. Ten class-stratified 75/25 development splits of 264 official training cuboids retain the official 132 test cuboids (66 sapwood/66 heartwood). Eighty opposite-class stress pairs maximize three-band cosine plus $0.10$ full-spectrum angle; both endpoints must be correct.

SpecTex uses per-spectrum centered/scaled bands at 450/550/650~nm. Ridge 1 predicts the standardized full spectrum; residual PCA tests $k\in\{0,2,4,8,16,32\}$ and the same scale grid. Validation bAcc$+0.1$ accuracy selects $k=32$ (35-D), followed by ridge 1. Each of 60 source arrays is shuffled with seed 20260620 and split by rounded 60\%/20\% counts, with the remainder for test: 107,181/35,728/35,731 spectra. This is within-source interpolation; source arrays span all partitions.
\begin{figure*}[!tp]
\centering
\includegraphics[width=0.97\textwidth]{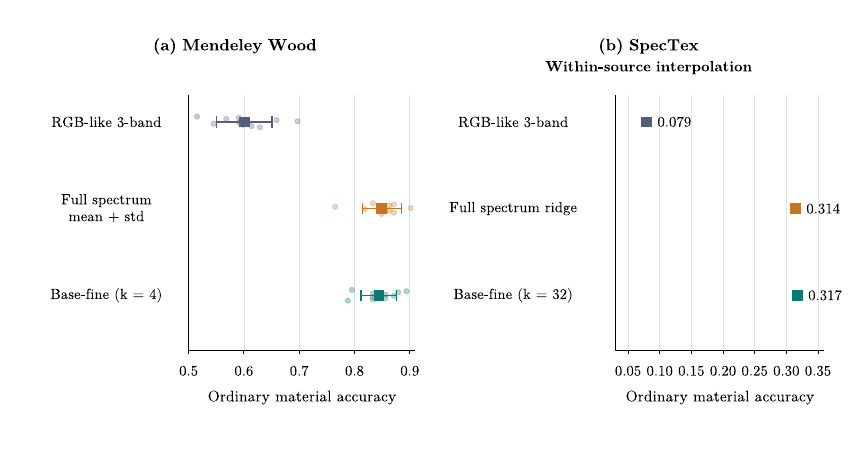}
\caption{Dataset-specific material augmentation. Width $k$ indexes the fine feature. Wood dots/squares/whiskers are runs/means/seed-split SD over ten development splits. SpecTex uses one within-source split across all 60 arrays. These are expanded material-readout features rather than native-width VLA states.}
\label{fig:supp-external-material}
\end{figure*}
Wood base-fine reaches bAcc $0.844$ versus full-spectrum mean+SD $0.849$, while resolving $0.975$ versus $0.782$ of constructed pairs. SpecTex accuracy is $0.079$ for the three-band base, $0.317$ for base-fine and $0.314$ for the full-spectrum ridge. Wood HyperSIGMA uniformly selects 120/320 bands, clips each cuboid by 1st/99th percentiles and resizes to $64^2$; 23 unavailable parameters include attention sampling offsets. WoodVIT \cite{woodvit} requires a complete NIR scene hierarchy for its official CNN split, unavailable in the examined subset.

\subsection{Sensor and Supervision Coverage}
SpectralWaste supplies synchronized imaging HSI/RGB, target masks and category labels. Wood, SpecTex and WoodVIT provide material evidence at specimen/image or spectrum level; driving resources HSI-Drive, HCV2/HyKo2 and Hyper-Drive provide scene segmentation with different RGB alignment \cite{hsidrive,hsiadapter,hyperdrive}. Erickson \emph{et al.} use point spectra \cite{robotmaterials}, while HSI-guided grasping supplies task-specific grasp supervision \cite{prismgrasp}. OmniVLA, MuseVLA, BridgeData V2, DROID and Open X-Embodiment supply language/trajectory supervision without imaging HSI \cite{omnivla,musevla,bridgedatav2,droid,openxembodiment}. These annotation differences motivate the target-grounded spectral-interface evaluation and distinguish it from trajectory-supervised policy evaluation.

%% file: tables/tab_material_state_mapping.tex
\begin{tabular}{@{}lcr@{}}
\toprule
\textbf{Category (ID)} & \textbf{State} & \textbf{Train/Val./Test} \\
\midrule
Film (1) & 1 & 203/67/66 \\
Basket (2) & 2 & 200/54/67 \\
Cardboard (3) & 0 & 35/15/10 \\
Video tape (4) & 3 & 397/117/130 \\
Filament (5) & 1 & 57/20/16 \\
Bag (6) & 1 & 500/169/172 \\
\bottomrule
\end{tabular}

%% file: tables/tab_session_spectral_confirmation.tex
\begin{tabular}{@{}lrrr@{}}
\toprule
\textbf{State} & $\boldsymbol{R^2\uparrow}$ & \textbf{Angle $\downarrow$} & \textbf{Category bAcc $\uparrow$} \\
\midrule
\rowcolor{proposedgray} \textbf{Continuous spectrum} & \textbf{0.802} & \textbf{0.020} & 0.861 \\
Category posterior & 0.337 & 0.038 & \textbf{0.882} \\
False-color base & 0.238 & 0.040 & 0.793 \\
\bottomrule
\end{tabular}

%% file: tables/tab_r7_coordinate_decomposition.tex
\begin{tabular}{lrrr}
\toprule
\textbf{Readout fitting} & \textbf{Raw} & \textbf{$G$} & \textbf{Injected} \\
\midrule
Raw-trained, fixed & 0.730 & 0.647 & 0.611 \\
$G$-trained, fixed & 0.609 & 0.704 & 0.620 \\
State-specific refit & 0.730 & 0.704 & 0.711 \\
\bottomrule
\end{tabular}

%% file: tables/tab_r7_head_occupancy.tex
\begin{tabular}{@{}lrrr@{}}
\toprule
\textbf{Quartile} & \textbf{Disagree (\%)} & \textbf{$T_q$/$C_q$ margin} & \textbf{$\Delta$} \\
\midrule
1 & 2.84 & 5.890/5.593 & -0.297 \\
2 & 3.51 & 4.910/4.857 & -0.053 \\
3 & 1.93 & 5.586/5.833 & +0.247 \\
4 & 3.21 & 6.157/6.818 & +0.661 \\
\bottomrule
\end{tabular}

%% file: tables/tab_base_shift.tex
\begin{tabular}{lrr}
\toprule
\textbf{Test rendering} & \textbf{Category bAcc} & \textbf{Pair} \\
\midrule
Low/mid/high & \textbf{0.929} & 0.827 \\
NIR edge & 0.918 & \textbf{0.852} \\
Random 3-band & \textbf{0.929} & 0.840 \\
Fixed sensor & 0.917 & 0.815 \\
Gain/offset & 0.913 & 0.840 \\
PCA & 0.918 & 0.827 \\
\bottomrule
\end{tabular}